\pdfoutput=1
\documentclass[preprint,12pt]{elsarticle}

\usepackage[utf8]{inputenc}
\usepackage[T1]{fontenc}
\usepackage{amsmath,amssymb}
\usepackage{graphicx}
\usepackage{booktabs}
\usepackage{tabularx}
\usepackage{multirow}
\usepackage{xcolor}
\usepackage{enumitem}
\usepackage{xurl}
\usepackage{hyperref}
\usepackage[capitalise,noabbrev]{cleveref}
\usepackage{lineno}

\graphicspath{{figures/}}

\newcommand{\simulean}{SimuLean}
\newcommand{\RQ}[1]{\textbf{RQ#1}}

\makeatletter
\def\ps@pprintTitle{\let\@oddhead\@empty\let\@evenhead\@empty\def\@oddfoot{\footnotesize\itshape Preprint\hfill\today}\let\@evenfoot\@oddfoot}
\makeatother

\begin{document}

\begin{frontmatter}

\title{Agentic discrete-event simulation for manufacturing system design: an
open 3D simulator, a benchmark and an industrial case}

\author[udc]{Javier Pernas-\'Alvarez\corref{cor1}}
\ead{javier.pernas2@udc.es}
\author[udc]{Diego Crespo-Pereira}
\cortext[cor1]{Corresponding author.}
\affiliation[udc]{organization={University of A Coru\~na, CITENI, Grupo Integrado de Ingenier\'ia},
            addressline={Campus Industrial de Ferrol, Campus de Esteiro s/n},
            city={Ferrol},
            postcode={15403},
            country={Spain}}

\begin{abstract}
The design of manufacturing systems relies on discrete-event simulation
(DES) models that experts still build in tool-specific interfaces. AI agents can now write
simulation code, but every model they write from scratch is new, unverified
software. We present \simulean{}, an open-source 3D DES simulator on the Unity
engine designed to be programmed by AI agents: its modelling semantics are
verified once, in the platform, and the agent builds and studies one model
through domain-level commands, in the 3D scene or headless; physically
simulated durations are calibrated for fast headless studies. We evaluated it on 18 manufacturing
modelling tasks, from single machines to published production lines. A frontier agent built valid models in 93 of 108 attempts
(86\%), in about two minutes each; on the benchmark's interface version,
two frontier agents built 105 of 108 models through domain commands alone, so
that each comes with a record of how it was built. Every residual failure was silent --- a
model that builds and runs but answers wrongly --- and only comparison with
reference models revealed them all. In an industrial case, one agent working alone rebuilt, verified and studied a
robotised welding workshop from a published study, in \simulean{} and in
SimPy. Comparing them with the published, expert-built model exposed an
error in it; against the corrected
model, the agent-built models agree within 1\%. One agent can already build
complete, realistic and traceable models for the design of manufacturing
systems; independent reference models are what make them trustworthy.
\end{abstract}

\begin{highlights}
\item Open 3D simulator lets AI agents build traceable models of manufacturing systems.
\item A frontier agent built valid models in 86\% of attempts, in about two minutes each.
\item Agent errors are silent: models run but answer wrongly; reference models catch them.
\item Agent-built models exposed an error in a published expert model of a real workshop.
\item Physically simulated transport calibrates travel models for fast headless studies.
\end{highlights}

\begin{keyword}
Discrete-event simulation \sep AI agents \sep Large language models \sep
Manufacturing systems \sep Model verification \sep Digital twin
\end{keyword}

\end{frontmatter}

\section{Introduction}\label{sec:intro}

Discrete-event simulation (DES) is the reference technique for analysing the
dynamic behaviour of manufacturing and logistics systems --- sizing capacity,
buffers and material handling, and evaluating changes before they are made
\citep{Jahangirian2010,Mourtzis2020}. With the digital twin its role has
grown: a DES model is often the component that predicts how a system will
behave under a decision not yet taken \citep{Kusiak2018,Tao2019,Lu2020}.
The way these models are built has changed far less. An expert translates the
system into a model through the graphical interface or scripting language of
a commercial package, each with its own vocabulary and hidden settings
\citep{Dagkakis2016}; modelling effort, licence cost and scarce expertise are
recurrent barriers, particularly for small and medium-sized enterprises
\citep{BicalhoHoch2022}, and open-source tools remove the licence cost but
not the need for expertise \citep{Lang2021}.

Large language models (LLMs), and in particular AI \emph{agents} that read
documentation, operate software tools and iterate on their own results over
many steps \citep{ren2025ai,Anthropic2024Agents}, could change who builds
simulation models and how. Recent work generates simulation code, usually
SimPy, from natural language \citep{Monks2025Unlocking,Chen2026SimLLM}, fills
templates or parameters of a commercial package
\citep{elbasheer2025natural,korth2026multiagentic}, or adapts a hand-written
model through a fixed chain of LLM roles with human checkpoints
\citep{schmitt2026llmdriven}. Expert opinion remains cautious: a 2025 panel
placed the capability to build a model and run the experiments that answer a
question one to two decades ahead \citep{Ehm2025LookFuture}.

Part of the difficulty lies in the tools. Simulators were designed for human
modellers: a model is spread between a drawing and settings hidden in
dialogues, runs are tied to the animation, and nothing checks automatically
that a change is correct; an agent reaches them through scripting interfaces
built for other purposes and receives little feedback it can act on. Writing
the model from scratch in a general-purpose language avoids the tool but
moves the burden: every feature a manufacturing study needs --- the layout,
vehicles travelling along aisles, shift calendars, failures counted on
operating time --- must be programmed and reviewed line by line, and the
result is a program a plant engineer cannot see. In both routes the most
dangerous errors are those that execute: models that run and look plausible
but are wrong \citep{Chen2026SimLLM,Fachada2026ODD}.

\begin{figure}[htbp]
\centering
\includegraphics[width=\textwidth,height=0.62\textheight,keepaspectratio]{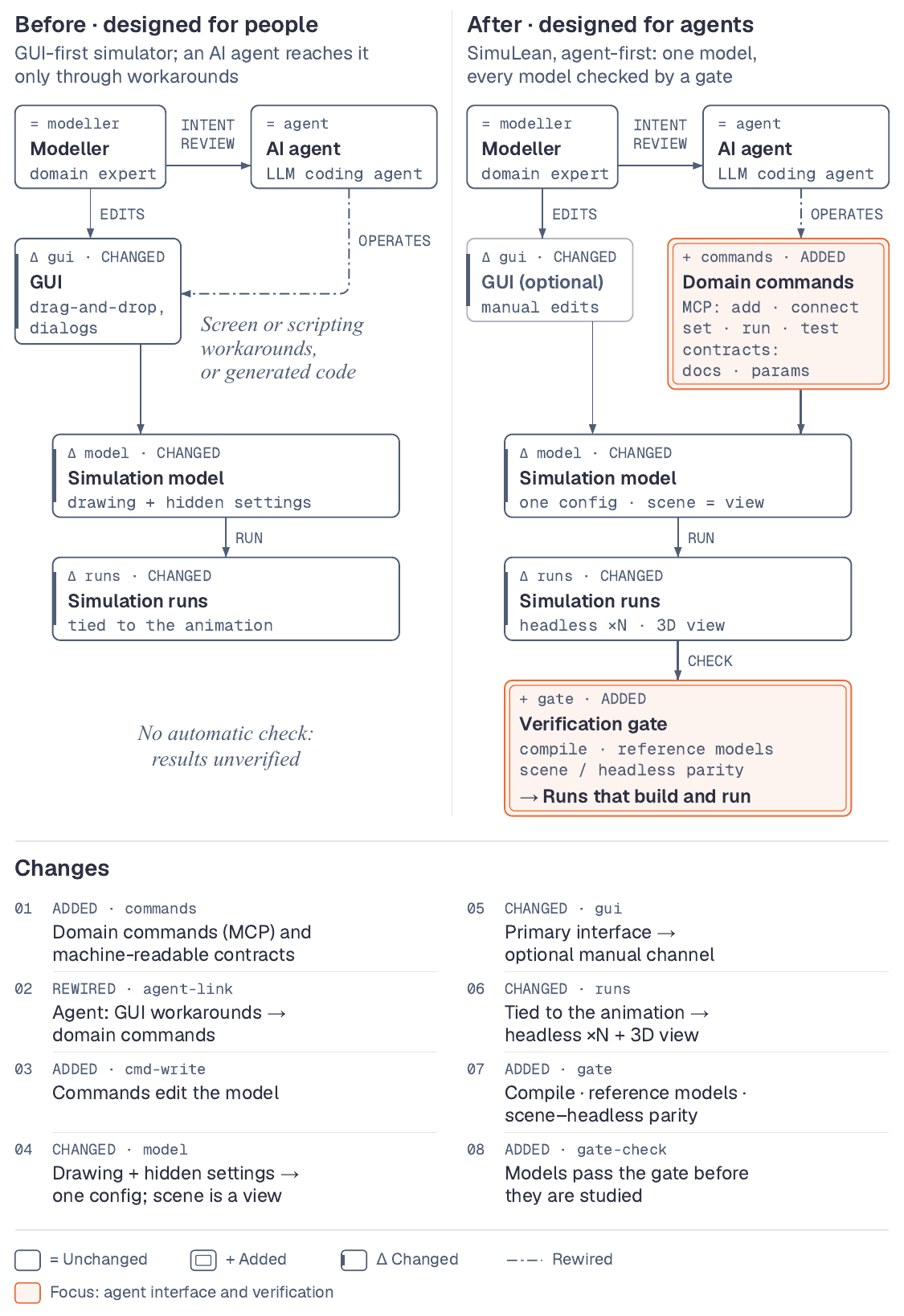}
\caption{A simulator designed for people versus one designed for AI agents.
Left: the agent reaches a GUI-first tool only through workarounds, the model is
split between the drawing and hidden settings, and nothing checks the results.
Right: in \simulean{} the agent works through domain commands and
machine-readable contracts on a single model whose scene is a view, runs it
headless or in 3D, and every model must build and run through a gate before
it is studied.
Below: the changes between the two. Diagram drawn with the assistance of Claude (Anthropic).}
\label{fig:designed-for-agents}
\end{figure}

We therefore design the simulator for the agent rather than adapt the agent
to an existing simulator. This paper presents \simulean{}, a free and
open-source 3D DES simulator on the Unity game engine designed to be
programmed by AI agents (\cref{fig:designed-for-agents}). Its modelling
semantics are verified once, in the platform; the agent works on one model
through domain-level commands, or code where they do not reach, and runs it in
the 3D scene or headless; manufacturing features --- aisle networks with AGVs
and forklifts, cranes, calendars, failures, teams, assembly and palletising
--- are configuration rather than code; and a duration can be ended by a
physical simulation in the scene and calibrated for headless studies. When
the agent works through the commands, the model it leaves behind is a
configuration of verified elements together with a complete record of how it
was built, rather than new code to be reviewed line by line. The claim we
test is one of sufficiency, not superiority over code: that such a simulator
lets one agent, working alone, build complete, realistic and traceable models
for the design of manufacturing systems, verify them and use them for
experiments and optimisation. Whether a simulator or code suits a given
project depends on what the project needs, and the evidence below is meant to
inform that choice.

We address four research questions:
\begin{enumerate}[label=\RQ{\arabic*}, leftmargin=*]
\item Can one AI agent, using \simulean{} as its primary modelling interface,
      build valid DES models of manufacturing systems from natural-language
      specifications, and at what effort?
\item Which errors do agents still make, and how can they be detected?
\item What does an agent-programmable 3D simulator add over an agent
      programming the model from scratch?
\item Does the simulator provide the computational performance and the
      physical fidelity of material flows that manufacturing studies need?
\end{enumerate}

We answer them with a benchmark of modelling tasks in five levels, with
validity criteria fixed before the runs, from a single machine to production lines built from published data,
including held-out tasks and run with two frontier models; with an arm in
which the same agent writes SimPy code from scratch; with an industrial case,
a robotised welding workshop rebuilt by one agent from a published study in
both \simulean{} and SimPy, whose comparison with the published model exposed
an error in that model and checked the agent's models against its corrected
version; and with performance and physics experiments. An independent
harness judges the model the agent leaves behind against reference models,
never the agent's own report.

The contributions of this paper are:
\begin{enumerate}[label=(\roman*), leftmargin=*]
\item \simulean{}, a free and open-source 3D DES simulator designed to be
      programmed by AI agents, in which the domain features of manufacturing
      systems are configuration rather than code and physically simulated
      durations are calibrated for headless studies, with the design
      principles it follows
      (\cref{sec:simulean});
\item a benchmark of manufacturing modelling tasks with reference models,
      behavioural validity criteria fixed before the runs and an evaluation
      harness that judges the model an agent leaves behind
      (\cref{sec:evaluation});
\item evidence on what frontier agents build with the simulator today, at
      what effort and cost, on the nature of their residual errors and how to
      detect them, and on what a 3D simulator adds over code written from
      scratch, including an industrial case in which agent-built models
      exposed an error in a published expert model and agree with its
      corrected version (\cref{sec:results,sec:case}).
\end{enumerate}

\Cref{sec:related} reviews related work and states the gap.
\Cref{sec:simulean} presents \simulean{} and the principles of a simulator
designed for agents. \Cref{sec:evaluation} describes the evaluation,
\cref{sec:results} reports the results for RQ1--RQ4, \cref{sec:case}
presents the industrial case, \cref{sec:discussion} discusses implications and
limitations, and \cref{sec:conclusions} concludes.

\section{Related work}\label{sec:related}

We review DES software and game engines, LLMs and agents that build
simulation models, and LLM agents under automated verification in other
engineering domains.

\subsection{DES software and game engines}
Industrial DES practice relies on commercial packages such as Plant
Simulation, FlexSim, Simio, AnyLogic and Arena, which combine graphical
modelling, 3D animation and experimentation, but whose closed source limits
inspection and extension and whose licences are a barrier for smaller
companies \citep{Dagkakis2016}. Open-source alternatives range from libraries
for programmers --- SimPy \citep{SimPyDocs}, salabim \citep{vanderHam2018},
Ciw \citep{Palmer2019Ciw} --- to manufacturing-oriented tools such as ManPy
\citep{Dagkakis2016ManPy} and JaamSim \citep{King2013}, which
\citet{Lang2021} consider a real alternative in production and logistics;
among them, only JaamSim and the earlier \simulean{} \citep{PernasAlvarez2023}
offer an integrated 3D modelling environment. All were designed for human
modellers and expose a model either as a drawing with settings in dialogues,
which an agent reaches only through a scripting interface, or as program
code, which an agent writes directly but only a programmer can review.

Game engines have been proposed as the basis of next-generation DES for their
physics and AI capabilities \citep{Turner2022NextGenDES} and are used for
manufacturing digital twins and extended reality
\citep{Scurati2025XRDES,Dodero2025DTSelection}. An external DES tool uses the
engine as a synchronised 3D view
\citep{Possik2021AnyLogicUnity,Andrade2025UnityFlexSim}, the DES logic runs
inside the engine \citep{PernasAlvarez2023,Fathi2023DESUnity}, or the engine
shadows physical equipment for virtual commissioning
\citep{Balla2023Game4Automation,Ullrich2024Omniverse}. Physics-based material
flow simulation captures effects that kinematic models miss
\citep{Glatt2021PhysicsDT,Cassano2025AGVUnity}, but we found no study that
quantifies its effect on DES results. LLMs have only begun
to reach these platforms --- for layout generation in Unity
\citep{wang2026llm}, robot commands in Isaac Sim \citep{Nambiar2025IsaacLLM}
and DEVS code with its Unity animation \citep{Cintra2025DEVSUnity} --- and
none lets an agent build, verify and experiment with a manufacturing DES
model inside the engine.

\subsection{LLMs and agents for simulation modelling}\label{sec:related-llm}
Work on LLMs for simulation modelling has passed through three generations.
In the \emph{chatbot} generation (2023--2024) a general-purpose model was
asked for simulation code of small queueing or inventory systems
\citep{jackson2024from,Frydenlund2024Modeler,Akhavan2024}.
\citet{Monks2025Unlocking} rebuilt two published healthcare models in SimPy
through 22 and 31 expert-guided iterations and reproduced the published
results in one of them.

\emph{Structured generation} (2024--2026) inserts an intermediate
representation between request and code: the DEVS formalism
\citep{CarreiraMunich2024DEVSCopilot}; 18 JSON templates filled by a local
open-weight model and mapped onto Plant Simulation components, with checks
that are structural by design \citep{elbasheer2025natural}; fine-tuned
models for supply-chain DES \citep{Chotaliya2025Generative} or SimPy
queueing tasks, for which executability proved an insufficient criterion
\citep{Chen2026SimLLM}; and schema-validated specifications from queueing
diagrams \citep{Monks2026Sketch2DES}.

\emph{Multi-agent workflows and tool-operating approaches} (2025--2026) are
closest to ours. \citet{schmitt2026llmdriven}, in this journal, adapt a
hand-written SimPy blueprint through a fixed chain of LLM roles with two
mandatory human checkpoints, validate the first model of each of two
industrial cases against a commercial tool and report 21 runs with their
cost; by their own definition theirs is a workflow rather than an agent
\citep{Anthropic2024Agents}. \citet{Kramer2026MAS} cover elicitation to
evaluation with a multi-agent system generating SimPy;
\citet{korth2026multiagentic} use four agents and MCP-registered tools to
parameterise a Plant Simulation model over seven tasks repeated twenty times
in one plant; and \citet{Jeong2025MCP} propose the Model Context Protocol
\citep{MCP2024} as the layer between LLM and simulator, with an M/M/1 proof
of concept. Other works use LLMs \emph{inside} a model built by people, e.g.\ as a
dispatching rule or an optimisation operator
\citep{dehghani2026simgpt,chen2026large,elbaz2026generative,xia2024llm,yuan2025chat}. Reviews name hallucination, reliability and
missing integration frameworks as the open problems
\citep{chen2025integrating,ren2025ai}, and the human-centric view of
Industry~5.0 favours the modeller in the loop
\citep{bajestani2026humanintheloop}.

Across the three generations, \emph{code written from scratch in SimPy} is the
most frequent target
\citep{Monks2025Unlocking,Chen2026SimLLM,schmitt2026llmdriven,Kramer2026MAS,Jeong2025MCP}:
the library is open, small and well documented, and a model is one program an
LLM can emit in one pass. It is the natural baseline for any claim about what
a simulator adds, yet no study compares an agent writing the model as code
with an agent operating a simulator on the same tasks. Correctness is
assessed on a ladder --- textual similarity, executability, consistency with
the specification, statistical equivalence with a reference --- and
executable models can still be wrong \citep{Chen2026SimLLM,Fachada2026ODD};
reported failures include API errors, plausible logic errors, context limits
and drift between model versions \citep{Monks2025Unlocking,Chen2026SimLLM}.
In the studies we read, success criteria are not fixed before the runs, no
held-out tasks guard against tuning the method to the evaluation, and
evaluations cover one or two cases or small queueing models.

\subsection{LLM agents under automated verification in engineering}
In PLC programming, adding a syntax checker, compiler and model checker raised
the success of LLM-generated code from 47\% to 72\% \citep{fakih2024llm4plc},
and verified agent pipelines report further gains
\citep{liu2024agents4plc,tu2026semaplc}; execution, simulation or geometric
feedback helps similarly for CAD \citep{alrashedy2025cadcodeverify}, Modelica
\citep{xiang2025modigen,wang2026pufibara} and optimisation models
\citep{ahmaditeshnizi2024optimus}. Without external feedback LLMs struggle to
correct themselves \citep{huang2024cannotselfcorrect,olausson2024selfrepair},
and test feedback drives the gains of reflective agents
\citep{shinn2023reflexion}. Agent evaluations should report repeated runs, error bars, cost and audited
test oracles
\citep{yao2024taubench,miller2024errorbars,kapoor2025agentsthatmatter,zhu2025abc}.
For simulation modelling it remains open whether an agent's errors are of the
kind a build-and-run check catches, and what a verifier must check to catch
the rest.

\subsection{Research gap}
\begin{figure}[htbp]
\centering
\includegraphics[width=0.85\textwidth,height=0.62\textheight,keepaspectratio]{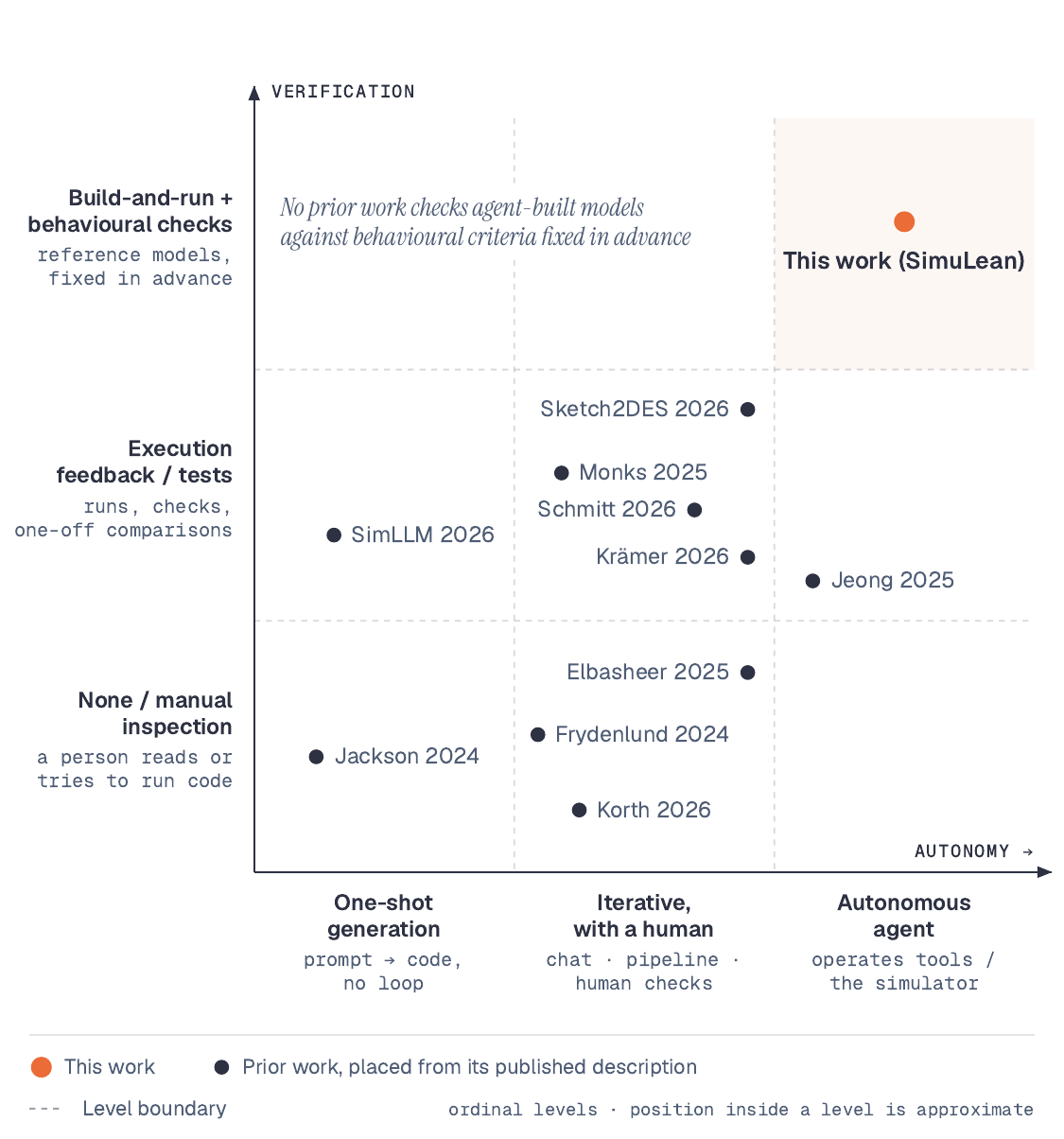}
\caption{Positioning of LLM-based simulation modelling work by agent autonomy
and automated verification (ordinal levels; positions within a level are
approximate, from each work's published description). To our knowledge, no
prior work checks the models an agent builds with build-and-run and
behavioural checks against criteria fixed before the runs. Diagram drawn with the assistance of Claude (Anthropic).}
\label{fig:positioning}
\end{figure}

\Cref{tab:gap} compares this work with the closest related work, and
\cref{fig:positioning} positions it by agent autonomy and automated
verification. Existing approaches generate stand-alone code or fill
templates or parameters of a commercial tool, run as fixed workflows rather
than agents that operate the simulator, verify through human review,
inspector LLMs or structural checks, and are evaluated on one or two cases
without criteria fixed in advance. To our knowledge, no prior work (i) offers
an open-source 3D simulator designed to be programmed by AI agents, with which
one general-purpose agent builds, verifies and experiments with
manufacturing DES models; (ii) evaluates such models against behavioural
validity criteria fixed before the runs, with reference models, repeated runs
and held-out tasks, and examines which errors survive a build-and-run check;
(iii) measures what such a simulator adds over an agent writing the same
models as code; or (iv) quantifies how physically simulated material flows
change the results of a manufacturing model and how they can be carried into
fast headless studies. We do not claim that agents need such a simulator to
succeed; the question is what it adds.

\begin{table}[htbp]
\centering
\caption{Closest related work on LLM-based construction of simulation models.
Verification: how the generated model is checked; Evaluation: as reported by
the authors. COTS: commercial tool; PS: Plant Simulation. Works in which the
LLM acts inside a model built by people are discussed in the text.}
\label{tab:gap}
\scriptsize
\setlength{\tabcolsep}{3pt}
\begin{tabularx}{\textwidth}{>{\raggedright\arraybackslash}p{2.1cm}
  >{\raggedright\arraybackslash}X >{\raggedright\arraybackslash}X
  >{\raggedright\arraybackslash}X >{\raggedright\arraybackslash}X}
\toprule
Work & Target & Agent role & Verification & Evaluation \\
\midrule
\citet{jackson2024from} & Python & One-shot & Manual review & 2 small models \\
\citet{Monks2025Unlocking} & SimPy & Iterative, expert-guided & Unit tests; vs.\ published results & 2 healthcare models \\
\citet{Chen2026SimLLM} & SimPy (fine-tuned 7B) & One-shot & Executability, format, consistency & 600 queueing tasks \\
\citet{Monks2026Sketch2DES} & DES code & Pipeline with reflection & Schema; one-off statistical comparison & 8 queueing diagrams \\
\citet{elbasheer2025natural} & PS (COTS) & Templates, local LLM & Schema only & 1 case; 199 requests; 1 LLM \\
\citet{schmitt2026llmdriven} & SimPy blueprint & Fixed LLM workflow; 2 human checkpoints & Inspector LLM; human review; one-off vs.\ COTS & 2 cases, 21 runs; cost \\
\citet{korth2026multiagentic} & PS (COTS) & 4-agent pipeline; parameterises & Plan checked in DES; model not verified & 7 tasks $\times$ 20, 1 plant \\
\citet{Kramer2026MAS} & SimPy & Multi-agent, end-to-end & Automatic validation & 1 case \\
\citet{Jeong2025MCP} & SimPy via MCP & Concept; proof of concept & Execution only & M/M/1 \\
\midrule
\textbf{This work} & \simulean{} (open, 3D, headless); SimPy arm & General agent operates the simulator & Build-and-run gate; behavioural validity vs.\ reference models, independent harness & 18 tasks (4 held out), $k=3$, 2 LLMs; SimPy arm; case; cost \\
\bottomrule
\end{tabularx}
\end{table}

\section{\simulean{}: a DES simulator designed to be programmed by agents}\label{sec:simulean}

\simulean{} is a DES simulator written in C\# and integrated with the Unity
game engine, released under the GNU General Public License v3.0 (GPL-3.0)
at \url{https://github.com/simulean/SimuLean}; version 1.0.0, the one described here, is archived on Zenodo \citep{SimuLeanZenodo}. It grew out of the engine of
\citet{PernasAlvarez2023} and has been redesigned so that an AI agent can be
its primary user: an agent, unlike a fixed workflow of LLM calls, decides
which tools to use and when \citep{Anthropic2024Agents}. The modeller owns
the question and the decision, the agent builds, verifies and experiments
with the model and reports its assumptions, and the simulator executes the
model and runs the automated checks; validation against the real system
stays with the modeller \citep{Law2024}. Manual editing remains a second
channel: the modeller watches the model as a plant in the 3D scene and can
open any element to change it, and the agent picks up the change in its next
step, since both act on the same model.

\subsection{Design principles}\label{sec:principles}
Designing for agents adds three goals to those of a tool for people: few
ways to be wrong, errors caught early and explained in terms the agent can
act on, and little effort per task. They lead to the six principles of
\cref{tab:principles} --- design goals, not established requirements;
whether an agent needs them is an empirical question (\cref{sec:results}).
An open simulator in a mainstream language also needs no simulation-specific
training data: what the agent does not know, it reads from the contracts or
the source.

\begin{table}[htbp]
\centering
\caption{Design principles of an agent-programmable simulator and how
\simulean{} implements them.}
\label{tab:principles}
\small
\begin{tabularx}{\textwidth}{@{}>{\raggedright\arraybackslash}p{3.6cm} >{\raggedright\arraybackslash}X@{}}
\toprule
Principle & In \simulean{} \\
\midrule
P1 Single source of truth & One configuration from which every run is built; the 3D scene is a view of it, not a second model \\
P2 Domain-level commands & The agent adds, connects, sets, runs and tests through the modelling vocabulary, not generic editor operations or arbitrary code \\
P3 Executable without the interface & Every model runs headless and in parallel, with the same results as the scene \\
P4 Deterministic reproducibility & One random stream per element and purpose; given a seed, results are identical in any mode, order or parallelism \\
P5 Machine-readable contracts & Conventions, the parameter catalog and every error are exposed in a form the agent reads before acting \\
P6 Automated verification gate & Every change to the simulator is checked against reference models; every model is built and run before it is accepted \\
\bottomrule
\end{tabularx}
\end{table}

\subsection{Architecture}\label{sec:arch-overview}
\begin{figure}[htbp]
\centering
\includegraphics[width=\textwidth,height=0.62\textheight,keepaspectratio]{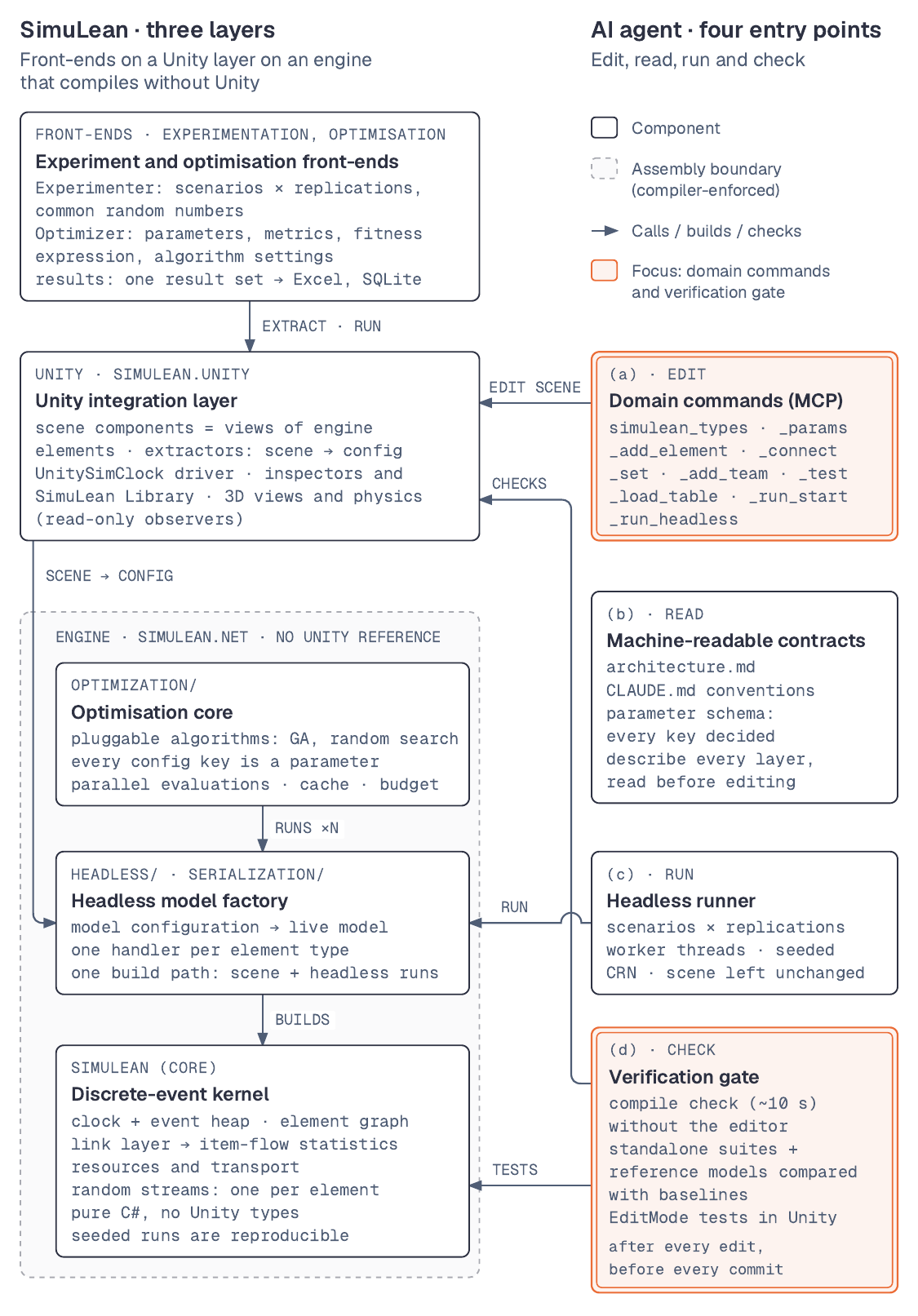}
\caption{\simulean{} architecture. Left: front-ends on the Unity integration
layer, on an engine assembly without any Unity reference. Right: the agent's
entry points --- (a) domain commands through the Model Context Protocol,
(b) machine-readable contracts, (c) the headless runner, (d) the verification
gate. Diagram drawn with the assistance of Claude (Anthropic).}
\label{fig:architecture}
\end{figure}

\simulean{} has three layers (\cref{fig:architecture}). The \emph{engine} is a
discrete-event kernel with its elements, link layer, statistics, resources,
transport, stochastics, model factory and optimisation core, compiled as an
assembly that cannot reference Unity --- a boundary enforced by the compiler
--- so every model also runs outside the game engine. The \emph{Unity integration layer} holds the
scene components that describe and display a model, the extractors that read
a scene, the element library and the commands exposed to the agent; the
\emph{front-ends} run experiments and optimisations (about 34\,000 and
35\,000 lines of C\# in the engine and the Unity layer).

The kernel uses next-event time advance; events scheduled for the same time
keep their scheduling order, so execution is deterministic. Timed work of an
element (processing, travel, set-up) is scheduled through the element, so a
failure, break or shift end pauses and resumes it. Statistics have exactly two
sources: item flow is recorded only by the link layer as items enter and
leave elements, and the state of resources and lists only by their own state
trackers, so no element type can report them inconsistently.

\begin{figure}[htbp]
\centering
\includegraphics[width=\textwidth,height=0.62\textheight,keepaspectratio]{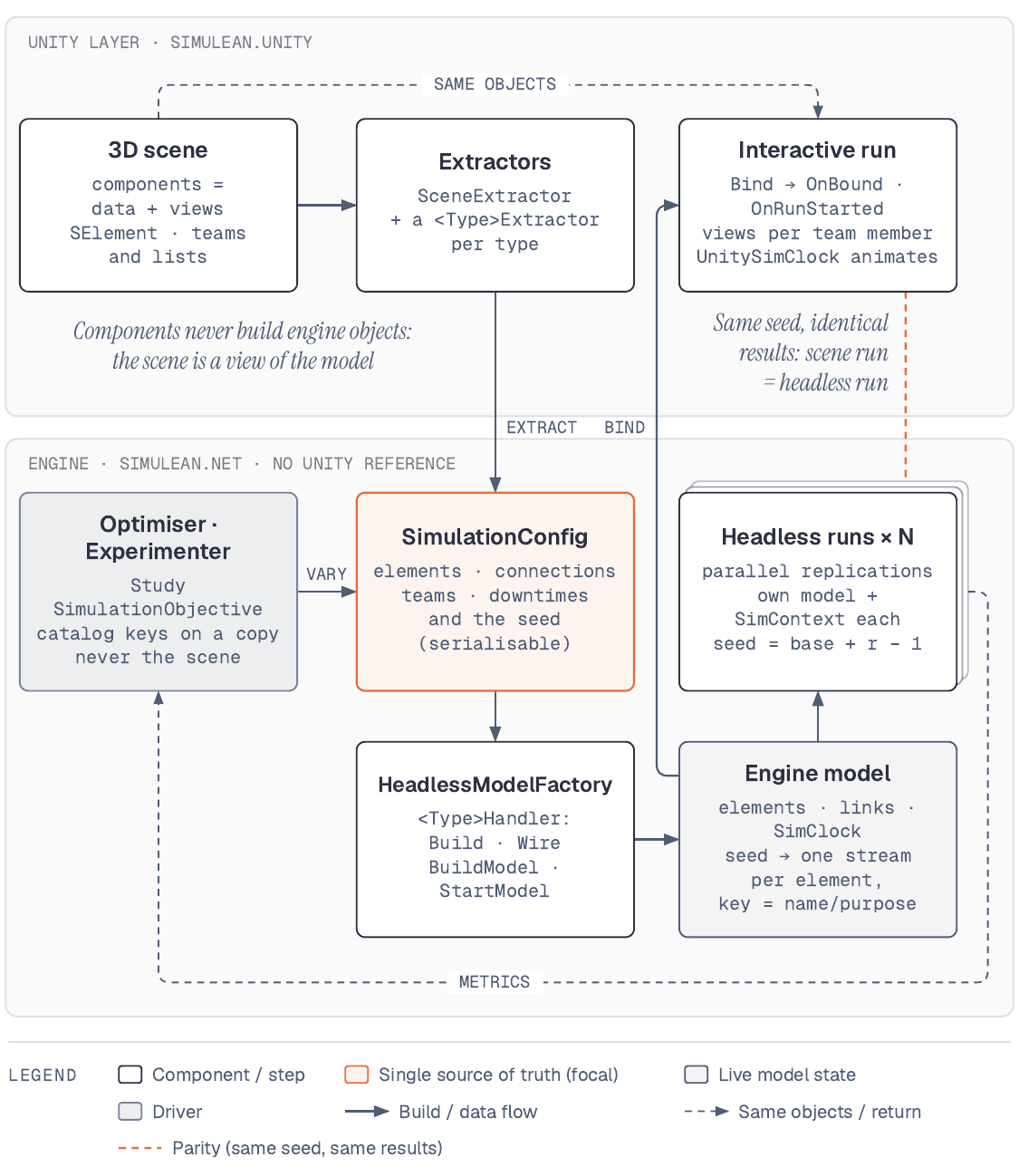}
\caption{One build path: the scene is extracted into a configuration from
which the engine builds the model; the scene components are bound to it as
views, or the same configuration runs headless in $N$ parallel replications.
The optimiser and experimenter vary the configuration, never the scene. Diagram drawn with the assistance of Claude (Anthropic).}
\label{fig:build-path}
\end{figure}

\paragraph{One build path (P1, P3)}\label{sec:build-path}
Scene components never build engine objects. When a run starts, the scene is
\emph{extracted} into a serialisable configuration --- elements, connections,
transport network, teams, lists, downtimes, reactions and input tables --- and
the engine builds the model from it alone, through one handler per element
type (\cref{fig:build-path}); the scene components are then bound to their
elements as views. A headless run executes the same configuration without the
scene, so interactive and headless runs are the same model built by the same
code. An agent therefore edits one representation and can run, test and
optimise every model without the 3D view. The view follows the model: a
resource team is stored as data and one visual object per member is created
when the run starts, so team size is a parameter like any other. Studies
currently run inside the editor process; a command-line runner that runs them
with the same engine outside it is being prepared for release, and in first
measurements it returned the same results while running long studies 14 to 25
times faster than the editor. Measured on the
industrial case (\cref{sec:case}), studies in the editor are about 10--14
times slower than the same engine outside it (about 10 times per replication;
50 replications took 4.3\,s outside on 8 threads, against about 3\,min per
150-run generation of the optimiser inside). The performance figures of
\cref{sec:rq4} are for the engine outside Unity.

\paragraph{Reproducibility (P4)}\label{sec:rng}
Every stochastic source draws from its own stream, seeded from the model seed
and a key formed by the owning element's name and the purpose of the draw
(inter-arrival, processing, failure, routing\ldots) through a fixed hash, not
from creation order. The same seed thus gives identical results, adding or
removing an unrelated element leaves the other draws unchanged, and parallel
runs equal sequential ones; replications use consecutive seeds, so scenarios
share common random numbers \citep{Law2024}. For an agent, a change in a
seeded result is evidence of a change in the model, never noise.

\paragraph{Studies and optimisation}\label{sec:catalog}
Every configuration key is a parameter. A catalog offers numbers, Booleans
and numeric arrays by type; text and list keys must be declared by their
handler, and the test suite fails if any key is left undecided, so the
catalog always answers ``what can I change in this model?''. The
\emph{Experimenter} runs a study that varies catalog parameters by path, as
scenarios or a full factorial, with replications and common random numbers,
and reports named metrics with confidence intervals; a study is checked
before it runs, and unknown names are reported with the closest valid ones.
The \emph{Optimizer} runs algorithms behind one interface that evaluates
candidates in parallel, caches them and enforces a budget; a genetic
algorithm for mixed spaces and random search are built in. Each evaluation
applies the candidate to a copy of the configuration, runs it headless and
computes a fitness expression in which constraints enter as penalties, so
a request such as the case study's (\cref{sec:case}) becomes a choice of
keys, bounds and a fitness expression.

\subsection{The agent interface}\label{sec:commands}\label{sec:agent-interface}
The modelling vocabulary is exposed to the agent as tools by a Model Context
Protocol server \citep{MCP2024} for the Unity editor (\cref{tab:commands}).
Every answer carries the errors and warnings logged while the command ran,
and unknown names come with the closest valid ones. \Cref{tab:commands}
lists the commands of the released version, refined after the experiments
of this paper: the commands marked $\dagger$ cover the steps for which agents
fell back on code. The benchmark used the earlier version v1, without them;
the industrial case was run on both versions (v1 and v2, \cref{sec:harness})
and is reported in \cref{sec:case2-runs}.

\begin{table}[htbp]
\centering
\caption{Domain-level commands exposed to the agent in the released version,
by family (prefix \texttt{simulean\_} omitted). $\dagger$: not in v1,
the version of the benchmark.}
\label{tab:commands}
\small
\begin{tabularx}{\textwidth}{@{}>{\raggedright\arraybackslash}p{4.4cm} >{\raggedright\arraybackslash}X@{}}
\toprule
Commands & What they do \\
\midrule
\texttt{types}, \texttt{params} & What can be placed; an element's fields and catalog keys \\
\texttt{add\_element}, \texttt{connect}, \texttt{set} & Place elements and tools, connect them (a track between transport nodes), set fields by name or path \\
\texttt{apply}$^\dagger$, \texttt{repeat}$^\dagger$, \texttt{invoke}$^\dagger$ & Many edits in one all-or-nothing call; copy a group of elements $N$ times with its internal connections; run an object's actions (e.g.\ fill a travel matrix) \\
\texttt{add\_team}, \texttt{load\_table} & Create resource teams; load input tables \\
\texttt{query}$^\dagger$, \texttt{fit}$^\dagger$ & Query or import into the project database (SQL, CSV, Excel); fit distributions to data and write the best into a field \\
\texttt{model}, \texttt{validate} & Read the whole model back as JSON; build it and check a study without running \\
\texttt{diagnose}$^\dagger$ & Run to a time or until the model stops and report stalls, deadlocks, what each element holds and waits for, and pending team requests \\
\texttt{run\_headless}; \texttt{run\_start}, \texttt{run\_status} & Run a study headless (scenarios, named metrics) and return each metric's mean and confidence half-width; the same in the background, with a wait until it ends \\
\texttt{optimize\_start}$^\dagger$, \texttt{optimize\_status}$^\dagger$ & Run the optimiser in the background (parameter ranges, metrics, fitness); return the best candidate re-run for every replication \\
\texttt{results}$^\dagger$ & Read, store and export result sets (CSV, Excel, SQLite, JSON) and draw charts \\
\texttt{test}, \texttt{test\_status} & Run the simulator's EditMode tests and report the outcome \\
\texttt{disconnect}, \texttt{remove}, \texttt{rename} & Undo mistakes; references to a removed object are cleared and listed \\
\texttt{place\_model}; \texttt{capture}$^\dagger$; \texttt{record\_*} & Place a library 3D model as decoration; render a framed picture of the model; record a video of a run \\
\bottomrule
\end{tabularx}
\end{table}

\paragraph{Contracts (P5)}
An architecture document and a conventions file, both checked against the
code by tests, state the rules the agent reads before acting (e.g.\ every
random draw needs its own stream). Invalid models fail at build time with
errors that name the element, field or table at fault.

\paragraph{Gate (P6)}
At the \emph{model level}, the one a modelling agent meets, the scene is
extracted, the model built, and a short headless run must end without an
error, a stall or a deadlock; a run monitor names each waiting element and
what it waits for. The gate checks that a model builds and runs, not that it
is right. At the \emph{developer level}, every edit of the simulator's own
code runs its regression suites (\cref{sec:simulean-verification}).

\subsection{Modelling elements}\label{sec:elements}
The element library is generated from the registered element types, so agent
and human see the same catalogue. \emph{Flow elements} include sources (from
distributions or schedule tables), queues and accumulation lanes,
multi-servers with parallel slots, delays, combiners that assemble a main
item with components in fixed quantities, assembly and picking stations,
stackers (palletisers) and storage systems with racks, access disciplines
and their own crane or vehicle team (an AS/RS); items leave by routing rules,
among them fixed probabilities per output. \emph{Resource teams} ---
operators, AGVs, forklifts, robots, cranes --- process or transport items,
dispatched by list queries such as the nearest free vehicle, and travel on a
\emph{transport network} of tracks along the aisles with optional traffic
control and deadlock resolution. One interruption mechanism serves failures
(in calendar or operating time), shifts with breaks, timetables and periodic
planned stops, for every element and resource. Lists support push and pull
logic, and reactions attach rules to element signals. Script-generated 3D
models of stations, vehicles and people show each element's state and
animate its work; they are views only and never affect the results.
Together, these elements cover the main kinds of construct that commercial
DES packages offer for manufacturing and logistics --- flow, resources and
teams, network transport, storage, assembly, calendars and failures,
experimentation and optimisation --- though not the breadth of their
specialised libraries.

\subsection{Discrete-event logic with physically simulated durations}\label{sec:physics}
In a DES model every activity has a duration known when it starts. \simulean{}
also lets a duration be open-ended and ended by a physical simulation in the
3D scene: the engine starts the activity, the physical component moves the
part or vehicle, and the engine resumes the activity when that component
reports that it has finished. The recorded durations are then fitted to a
calibrated model that headless studies use. The mechanism is implemented for
transport and designed to extend to other processes whose duration depends
on physics, such as handling, falling or placing material. For transport,
travel can be modelled with increasing fidelity: (T0) a defined duration;
(T1) network distance over speed, a distance matrix or trapezoidal axis
profiles, optionally with traffic control; (T2) A* paths on a walkable grid;
and (T3) movement simulated by Unity's physics and navigation, with
collisions and congestion emerging from the scene; the recorded T3 moves fit
a calibrated T1 model (per-pair means or a regression on distance and load).
Physics runs depend on frame timing, so they are interactive only and outside
the reproducibility guarantee; headless runs and optimisation use the
calibrated model, and an uncalibrated physics team cannot be exported.
Physics-based emulation and virtual-commissioning tools exist
\citep{Balla2023Game4Automation,Ullrich2024Omniverse}, but they typically
target bounded work cells; \simulean{} couples physics with plant-scale DES
models and carries it, through calibration, into fast headless studies.

\subsection{\simulean{} among DES tools}\label{sec:capabilities}
\Cref{tab:capabilities} compares \simulean{} with two open-source libraries
(SimPy, salabim), an open-source 3D simulator (JaamSim) and three commercial
packages, from their documentation and published work. It records whether a
capability is offered and in what form, not how well it works: libraries can
program almost anything, and commercial packages add much by scripting. That
difference --- configured versus programmed and reviewed as code --- is what
RQ3 examines.

\begin{table}[htbp]
\centering
\caption{Capabilities relevant to agent-built manufacturing models.
--: not found in the documentation consulted (not a statement that it cannot
be done, e.g.\ by programming); Plant Sim.: Plant Simulation.
\textsuperscript{a}Concept only: an LLM wrote SimPy code from MCP-compatible
context \citep{Jeong2025MCP}.
\textsuperscript{b}Community or third-party MCP servers (e.g.\ a commercial
assistant that drives FlexSim from AI coding tools), not official releases
\citep{FlexSimMCPCommunity,FlexSimMCPIdea}.
\textsuperscript{c}No native MCP server; an AI copilot is offered in Plant
Simulation~X, and a research prototype drives Plant Simulation through COM
\citep{korth2026multiagentic}.
\textsuperscript{d}Read-only JSON description of the whole model.
\textsuperscript{e}Not confirmed in the public documentation consulted.}
\label{tab:capabilities}
\scriptsize
\setlength{\tabcolsep}{2.5pt}
\renewcommand{\arraystretch}{1.25}
\begin{tabularx}{\textwidth}{@{}>{\raggedright\arraybackslash}p{2.0cm}
  *{7}{>{\raggedright\arraybackslash}X}@{}}
\toprule
 & \simulean{} & SimPy & salabim & JaamSim & FlexSim & AnyLogic & Plant Sim. \\
\midrule
Open source; free & yes (GPL-3.0) & yes (MIT) & yes (MIT) & yes (Apache 2.0) & no; free student ed.\ (Autodesk) & no; free learning ed. & no; free student ed. \\
3D layout, animation & yes & no & yes (2D, 3D) & yes & yes & yes & yes \\
Random stream per element & yes, default (by name) & by scripting & optional, per distrib. & yes (seed per distrib.) & yes (per object) & by scripting & n.c.\textsuperscript{e} \\
Agent interface (MCP) & yes, native & --\textsuperscript{a} & -- & -- & community / third party\textsuperscript{b} & community\textsuperscript{b} & no native\textsuperscript{c} \\
Machine-readable model & partial\textsuperscript{d} & code & code & yes (text) & yes (XML) & yes (XML) & via interfaces \\
Experimenter / optimiser & yes / yes (GA, random; pluggable algorithms) & -- / -- & -- / -- & yes / -- & yes / yes (OptQuest add-on) & yes / yes (OptQuest) & yes / yes (GA) \\
Physics-based simulation & yes (ends DES durations; calibrated headless) & -- & -- & -- & -- & -- & -- \\
Scripting / API & C\# & Python & Python & expressions; Java & FlexScript & Java & SimTalk; COM \\
\midrule
Sources & this paper & \citep{SimPyDocs} & \citep{vanderHam2018,SalabimDocs} & \citep{King2013,JaamSimManual2021} & \citep{FlexSimManual2016} & \citep{AnyLogicProduct,AnyLogicHelp} & \citep{PlantSimProduct,PlantSim15Blog} \\
\bottomrule
\end{tabularx}
\end{table}

\subsection{Verification of the simulator}\label{sec:simulean-verification}
The simulator's semantics are verified once, in the platform, by a regression
net that runs after every change to its code: 20 standalone test suites,
among them 43 single-purpose reference models whose per-element statistics
must match stored baselines within a relative tolerance of $10^{-6}$, and 212
EditMode tests in Unity, all of which passed at the version evaluated. Scene
and headless runs of the same model and seed agree exactly on throughput and
within $10^{-12}$ on the other statistics. The few faults of the simulator
that came to light while agents used it during this work were fixed in the
released version. Against queueing theory, 12 of 15
closed-form values of the benchmark's reference models lie inside the 95\%
confidence interval of the simulated estimate over 100 replications; the
other three share one sample of arrivals and lie within 0.5\% of theory.
Verifying the simulator is not validating a model built with it: whether a
model represents its system is checked separately (\cref{sec:evaluation}).

\section{Evaluation}\label{sec:evaluation}

The evaluation answers RQ1 and RQ2 with a benchmark of modelling tasks and
an industrial case, RQ3 with the same agent programming models in SimPy, and
RQ4 with performance and physics experiments (\cref{fig:design}). Tasks,
reference models and validity criteria were fixed before the main runs.

\subsection{Benchmark}\label{sec:benchmark}
The benchmark has 18 tasks on five levels: L1 basic flow (a single machine,
a serial line with blocking, parallel machines), L2 resources and time
(shared operators with shifts and breaks, failures and planned maintenance,
assembly), L3 material handling (AGVs on a path network, rack storage and
picking, job priorities), L4 change and decide (modify a given model, run an
experiment, allocate buffer space) and L5 industrial lines, two real lines
from the public FlowLineLib library \citep{Weiss2019} given as its data.
Four further tasks (H1--H4) were held out: they were committed with a content
hash and not used while designing the agent interface. Each task is a short
natural-language specification that names the elements whose KPIs are
measured; the box shows what the agent receives for T1.2, besides a fixed
system prompt. The tasks are small compared with real plants.

\begin{center}
\fbox{\begin{minipage}{0.94\textwidth}\small
\textbf{T1.2 Serial line with finite buffers.} A four-machine line
\texttt{M1} $\rightarrow$ \texttt{B1} $\rightarrow$ \texttt{M2}
$\rightarrow$ \texttt{B2} $\rightarrow$ \texttt{M3} $\rightarrow$
\texttt{B3} $\rightarrow$ \texttt{M4} $\rightarrow$ \texttt{Exit}. Raw
material is always available in front of \texttt{M1}. Each buffer
\texttt{B1}--\texttt{B3} holds at most 3 parts; when a buffer is full, the
machine before it keeps the finished part and is blocked until space frees
up. Processing times are exponential, with means 3.3\,min (\texttt{M1}),
4.3\,min (\texttt{M2}), 3.0\,min (\texttt{M3}) and 3.7\,min (\texttt{M4}).
Report the throughput per hour and the utilisation of each machine.
\end{minipage}}
\end{center}

\subsection{Agents, interface versions and harness}\label{sec:harness}
The main model is Claude Opus~5.5; the second is Claude Sonnet~5.5; both ran
all 18 tasks on v1. Both ran in Claude Code (pinned
version), with the \simulean{} commands (\cref{sec:commands}) and the
editor's generic tools, among them one that evaluates C\# code in the
editor, read-only file tools, no shell and no file editing.
Each task was run $k=3$ times per condition, each run in a fresh
non-interactive session with the task text as the only message, from a clean
scene, with no human in the loop and limits of 30\,min, 150 tool calls and
$5\times10^{6}$ tokens. Opus and Sonnet ran on two versions of the agent
interface, v0 (Sonnet at L1--L3 only) and v1, where v1 adds errors and warnings to every answer,
checks studies before they run, and adds commands that read and validate the
whole model and undo mistakes, with the engine unchanged. We call v1 the
version used in the benchmark, and v2 the released version, with the refined
domain commands of \cref{tab:commands} and an agent documentation; v2 was
used only in the industrial case. Two ablations on v1
(Opus, L1--L3) removed the build-and-run gate (C-NoGate) or the \simulean{}
commands, leaving generic editor tools and code evaluation (C-Generic). These
six sets of runs --- Opus and Sonnet on v0 and v1, and the two ablations ---
form the main block of 243 attempts. Effort is the agent's wall time, tokens
and tool calls; cost is Claude Code's API-equivalent estimate, since the runs
used a subscription. For an indicative manual baseline, the two authors, both familiar with the
simulator, built T1.2 and T2.3 by hand.

\subsection{Validity criterion}
The harness extracts and runs the model each attempt leaves behind, never
trusting the agent's report, and compares it with a reference model written
by the authors. A model is \emph{valid} if it behaves like the reference:
every target KPI (throughputs, utilisations, mean contents and waits) must
agree within a margin of $\pm5\%$ for throughput and utilisation and
$\pm15\%$ for contents and waiting times. Statistically, the 90\% confidence
interval of the relative difference of the means must lie inside the margin.
The interval is Welch's, which compares two means without assuming equal
variances \citep{Welch1947}, from 100 replications of each model; requiring it
to lie inside the margin is the two one-sided tests procedure (TOST)
\citep{Schuirmann1987}, which declares two models equivalent only when the
difference is shown to be both above the lower and below the upper margin.
The criterion was calibrated so that a correct model passes with probability
at least 0.95. Every attempt is classified by the first step
it fails: no model (none, or a run ended by a limit), build failure, gate
failure, \emph{silent error} (a model that builds, runs and passes the gate
but is not valid) or valid. Rates carry Wilson 95\% intervals
\citep{Wilson1927}, an interval for a proportion that stays within $[0,1]$
and holds for small samples; paired comparisons use McNemar's exact test
\citep{McNemar1947}, which compares two success rates on the same tasks using
only the pairs whose outcomes differ.

\subsection{SimPy comparison}\label{sec:simpy-design}
The same agent (Opus, same limits) built six of the tasks --- T1.2, T2.1,
T2.3, T3.1, T4.1 and T5.1, one or two per level --- from scratch in
SimPy~4.1.2 \citep{SimPyDocs}, $k=3$, with a shell restricted to
\texttt{python}. A fixed contract followed the task text: deliver a
\texttt{model.py} whose \texttt{run(seed, horizon, warmup)} returns the
target KPIs, each with its measurement definition. Seeds, margins and
reference values are those of the \simulean{} v1 runs, paired by task and
repetition.

\subsection{Industrial case}\label{sec:case-design}
A request of the size a plant engineer would make --- rebuild a published
industrial workshop, verify it, run an experiment and an optimisation, and
report (\cref{sec:case}) --- was given to Opus with the benchmark harness and
raised limits (400 tool calls and $40\times10^{6}$ tokens per session), three
times on v1 and three times on v2, which adds to v1 the commands marked
$\dagger$ in \cref{tab:commands} and an agent documentation (workflow,
command reference, error catalogue). The specification, limits,
model and fixed resume message were the same, and nobody intervened in the
model. The same agent also built the workshop three times in SimPy with the
same specification and the SimPy contract. The harness re-ran each exported
model with 50 replications and compared it with the published model, which
was corrected during this work; the final runs in both tools used the
specification with the corrected rule (\cref{sec:case2,sec:case2-val}), so
that each of the three arms ran three times on each of the two forms of the
specification (18 runs). With three runs per version and specification, the
case shows what one agent does at plant scale; it does not
test a difference between versions.

\subsection{Performance and physics}
An M/M/1 queue, tandem lines of 5 to 200 stations and the hold model
\citep{Jones1986} were run headless on one thread (events per second) and
compared with statistically equivalent SimPy and salabim
\citep{vanderHam2018} models by wall time per replication, on a laptop
(Intel Core i7-12650H). For physical material flows (\cref{sec:physics}),
a fleet of 1 to 4 AGVs serves a three-machine cell in a wide two-way and a
narrow single-lane layout (2 to 4 AGVs in the narrow one) under four travel
models: constant times (T0), an aisle network (T1), an A$^{*}$ grid (T2) and
Unity navigation with physics (T3, interactive scene only, five seeds), plus
T1 calibrated from the T3 moves (T1-cal); the headless variants use 100
replications. The fleet is chosen as the smallest that meets two targets
fixed before the T3 runs: a throughput of at least 38 parts per hour and a
mean flow time of at most 720\,s.

\begin{figure}[htbp]
\centering
\includegraphics[width=\textwidth,height=0.62\textheight,keepaspectratio]{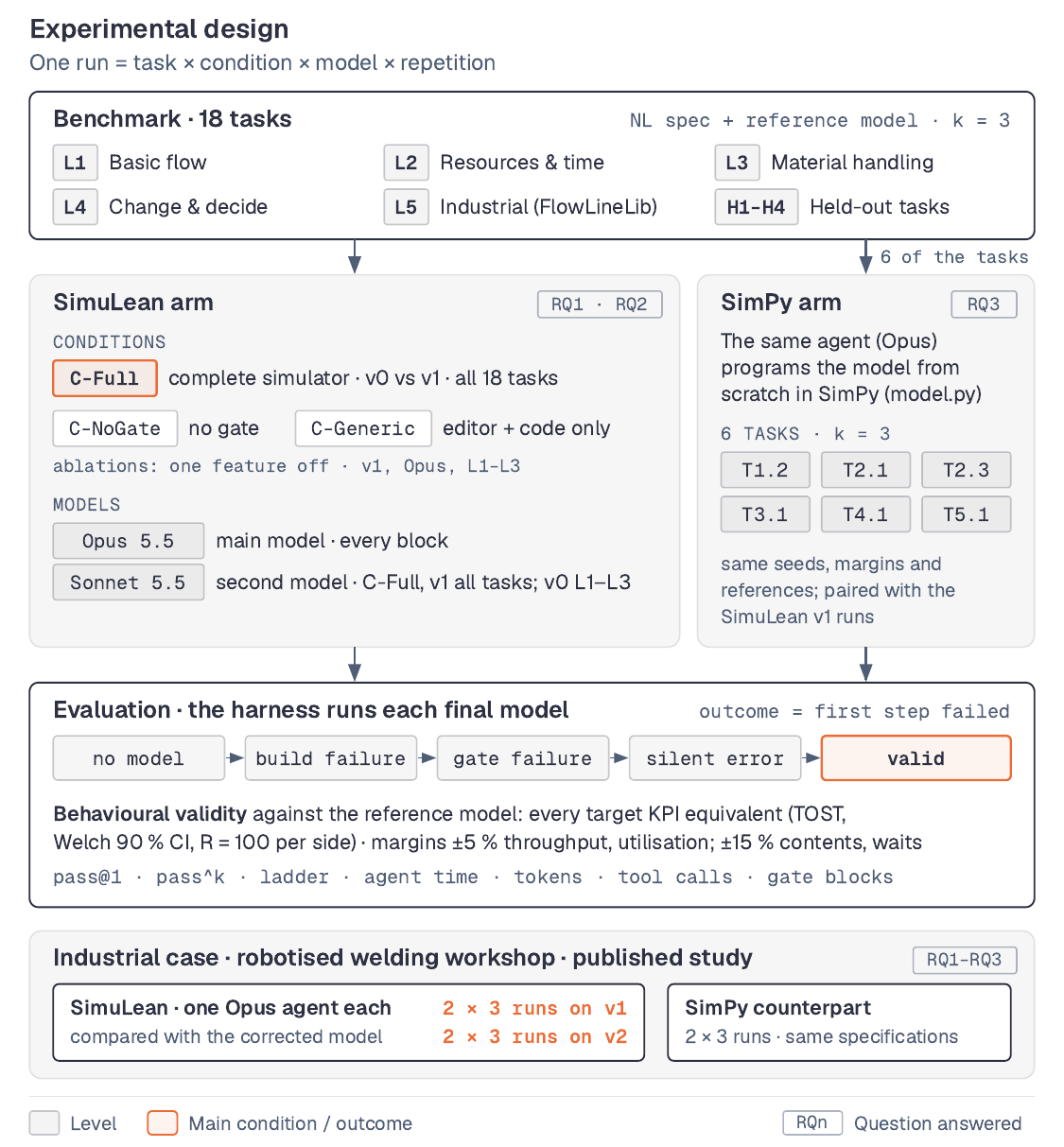}
\caption{Evaluation design. The benchmark (L1--L5 and held-out tasks, $k=3$)
runs in the \simulean{} arm (C-Full on two interface versions, ablations
C-NoGate and C-Generic; Opus and Sonnet) and, for six tasks, in the SimPy
arm. The harness runs every final model and classifies it by the first step
it fails, with validity judged against a reference model. The industrial
case is built three times on each of two interface versions and three times
in SimPy, and validated against the corrected published model. Diagram drawn with the assistance of Claude (Anthropic).}
\label{fig:design}
\end{figure}

\section{Results}\label{sec:results}

\subsection{RQ1: validity and effort of agent-built models}\label{sec:rq1}

\emph{One agent working alone built a valid model in most attempts and at
every level, typically in about two minutes and for about one US dollar.}

\paragraph{Validity}
Claude Opus~5.5 built a valid model in 93 of 108 attempts over the two
interface versions (0.86 [0.78, 0.91]; \cref{tab:rq1}), 77 of 84 on the 14
main tasks. Every attempt at L1, L3 and L5 was valid, including the twelve at
the two FlowLineLib lines; the misses concentrate in one L2 task (T2.1), two
held-out tasks (H1, H4) and two runs at L4 that ended at the token cap. Claude
Sonnet~5.5 was valid in 43 of 54 attempts on v1 over all 18 tasks (25 of 27
at L1--L3), against 46 of 54 for Opus on the same tasks (paired by task and
repetition, exact McNemar, two-sided, $p=0.45$); its misses concentrate in
T2.1, two L4 tasks (T4.2, T4.3), H2 and H4. The two versions gave
similar results, with no detectable difference (Opus 47 and 46 of 54, McNemar
$p=0.875$; Sonnet 23 and 25 of 27), and so did the ablations on v1: 23/27 without the gate
(C-NoGate) and 25/27 with generic tools only (C-Generic), against 24/27 for
C-Full; C-Generic halved the tool calls (median 53 to 28) because the agent
wrote larger units of work as code. In C-Full, by contrast, 53 of the 54
Opus attempts and 52 of the 54 Sonnet attempts on v1 built the model through domain
commands alone (median 49 commands for Opus and 46 for Sonnet, no line of
code), so each model comes with
a complete record of how it was built; every C-Generic attempt wrote code
(median 11 blocks and 51 lines). This is a trade-off rather than a gain in
validity: fewer calls, against a model whose construction can be read step by
step. C-Generic's code, moreover, still calls
\simulean{}'s verified engine, unlike code written from scratch, whose far
end the SimPy arm (\cref{sec:rq3}) and the industrial case
(\cref{sec:case-simpy}) show. No attempt asked a question or was helped
by a person. Over the main block of 243 attempts, 207 were valid. The
industrial case extends the question to plant scale (\cref{sec:case}).

\begin{table}[htbp]
\centering
\caption{Valid models by level and model, C-Full, both interface versions
pooled (valid / attempts, Wilson 95\% CI). Sonnet: L4, L5 and held-out on
v1 only, so its totals pool 54 attempts at L1--L3 with 27 at L4--H4.}
\label{tab:rq1}
\small
\begin{tabularx}{\textwidth}{@{}X c c c c@{}}
\toprule
 & \multicolumn{2}{c}{Claude Opus 5.5} & \multicolumn{2}{c}{Claude Sonnet 5.5} \\
\cmidrule(lr){2-3}\cmidrule(lr){4-5}
Level & Valid & 95\% CI & Valid & 95\% CI \\
\midrule
L1 Basic flow           & 18/18 & [0.82, 1.00] & 18/18 & [0.82, 1.00] \\
L2 Resources \& time    & 13/18 & [0.49, 0.88] & 13/18 & [0.49, 0.88] \\
L3 Material handling    & 18/18 & [0.82, 1.00] & 17/18 & [0.74, 0.99] \\
L4 Change \& decide     & 16/18 & [0.67, 0.97] & 4/9   & [0.19, 0.73] \\
L5 Industrial lines     & 12/12 & [0.76, 1.00] & 6/6   & [0.61, 1.00] \\
Held-out H1--H4         & 16/24 & [0.47, 0.82] & 8/12  & [0.39, 0.86] \\
\midrule
L1--L3                  & 49/54 & [0.80, 0.96] & 48/54 & [0.78, 0.95] \\
Main tasks (L1--L5)     & 77/84 & [0.84, 0.96] & 58/69 & [0.74, 0.91] \\
All 18 tasks            & 93/108 & [0.78, 0.91] & 66/81 & [0.72, 0.88] \\
\bottomrule
\end{tabularx}
\end{table}

\begin{figure}[htbp]
\centering
\includegraphics[width=0.8\textwidth]{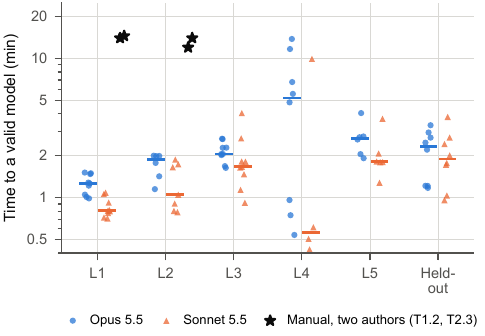}
\caption{Agent time to a valid model by task level (v1, C-Full; log scale).
Points are valid attempts, bars the medians. Stars:
the manual baseline, the two authors building T1.2 (L1) and T2.3 (L2) by
hand ($n=2$, familiar with the simulator, hence optimistic).}
\label{fig:effort}
\end{figure}

\paragraph{Time}
On v1 the median time to a valid model was 2.0\,min for Opus (46 valid of
54 attempts) and 1.5\,min for Sonnet (43 of 54); 42 of the 46 Opus models
took less than five minutes (\cref{fig:effort}). The medians grow from about
1.3\,min at L1 to 2--3\,min at L2, L3, L5 and the held-out tasks; L4 is the
widest, from half a minute to add one machine to a given model to about
14\,min to search buffer allocations. In the manual baseline, the two authors built T1.2 and T2.3 by hand in about
14--14.5 and 12--14\,min to a valid model (one parameter-entry error
corrected in each of the first author's models: an exponential mean typed into
the rate field), against agent medians of 1.5 and 1.4\,min on the same tasks;
with two modellers who know the simulator, this indicates an order of
magnitude only.

\begin{figure}[htbp]
\centering
\includegraphics[width=\textwidth]{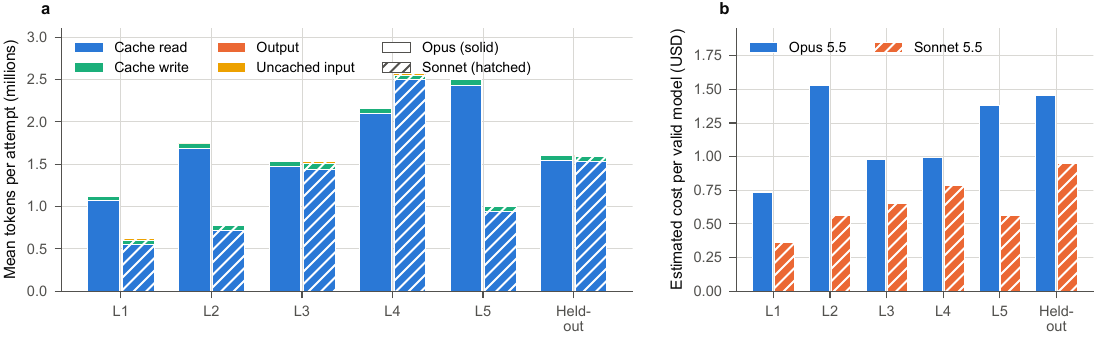}
\caption{Effort in tokens and cost (v1, C-Full). (a) Mean tokens per attempt
by type --- cache read, cache write, output and uncached input --- by level,
Opus solid and Sonnet hatched. (b) Estimated cost per valid model: Claude
Code's API-equivalent estimate for all attempts at a level divided by the
valid models (the runs used a subscription; no estimate exists for a run ended by the token cap).}
\label{fig:tokens}
\end{figure}

\paragraph{Tokens and cost}
An Opus attempt processed a mean of $1.74\times10^{6}$ tokens, of which 96\%
were cache reads --- the context of the editor's tools and earlier answers,
re-read on every turn --- and 0.6\% output (\cref{fig:tokens}a); a Sonnet
attempt processed $1.38\times10^{6}$. Claude Code estimated the API-equivalent
cost of the 54 Opus attempts at 52.5\,USD, or 1.14\,USD per valid model, from
0.74\,USD at L1 to 1.53\,USD at L2; for Sonnet the estimate is 0.63\,USD per
valid model (\cref{fig:tokens}b); Claude Code gives no estimate for a run
ended by the token cap, so those runs are not included. Six of the 243 runs ended at the token cap,
four of them at T4.2, where the agent polled the status of a long study until
the cap (v1 cannot wait on a running study); none reached the time or
tool-call limit.

\subsection{RQ2: the errors agents still make and how to detect them}\label{sec:rq2}

\emph{Every attempt that failed and was not ended by a limit was a silent
error: a model that builds, runs and passes the gate but answers the question
wrongly (\cref{tab:failures}). The build-and-run gate caught none;
behavioural checks on the model caught about half, but not the errors in reading
the specification.}

\begin{table}[htbp]
\centering
\caption{Outcomes of the 243 attempts of the main block by class (counts). No
attempt failed at the build or gate step.}
\label{tab:failures}
\small
\setlength{\tabcolsep}{4pt}
\begin{tabularx}{\textwidth}{@{}X c c c c c c c@{}}
\toprule
 & \multicolumn{2}{c}{Opus C-Full} & \multicolumn{2}{c}{Sonnet C-Full} & NoGate & Generic & \\
\cmidrule(lr){2-3}\cmidrule(lr){4-5}
Outcome / class & v0 & v1 & v0 & v1 & v1 & v1 & Total \\
\midrule
Attempts                              & 54 & 54 & 27 & 54 & 27 & 27 & 243 \\
Valid                                 & 47 & 46 & 23 & 43 & 23 & 25 & 207 \\
Build or gate failure                 & 0  & 0  & 0  & 0  & 0  & 0  & 0 \\
Silent: operator busy in break (T2.1) & 2  & 3  & 3  & 2  & 3  & 2  & 15 \\
Silent: ambiguous spec.\ (H4)         & 3  & 3  & -- & 3  & -- & -- & 9 \\
Silent: rule added (H1)               & 1  & 1  & -- & 0  & -- & -- & 2 \\
Silent: time unit                     & 0  & 0  & 0  & 2  & 1  & 0  & 3 \\
Silent: forklift handling (H2)        & 0  & 0  & -- & 1  & -- & -- & 1 \\
No model: token cap                   & 1  & 1  & 1  & 3  & 0  & 0  & 6 \\
\bottomrule
\end{tabularx}
\end{table}

\paragraph{Silent errors}
Fifteen of the 30 silent errors are the same mistake on T2.1: the agent
stopped the two stations for the operator's break but not the operator, so a
part paused by the break kept its operator and the engine booked the break as
busy time (utilisation 0.94 against 0.88), while throughput and queue
contents were correct. It occurred with both models, on both versions and in
every condition. Nine are H4, whose prompt says ``demand has doubled'' for a
start model that already has the doubled demand: all nine attempts (Opus and
Sonnet) doubled it again. Two (H1) added a priority rule that the
specification does not contain; one left the clock in its default unit where
the task gives minutes, and two (T4.3) read the minutes of the task as hours
--- both found the reference's best buffer allocation, with a throughput 60
times too low; one (H2) modelled the forklift's handling differently from the
specification.

\begin{figure}[htbp]
\centering
\includegraphics[width=0.6\textwidth]{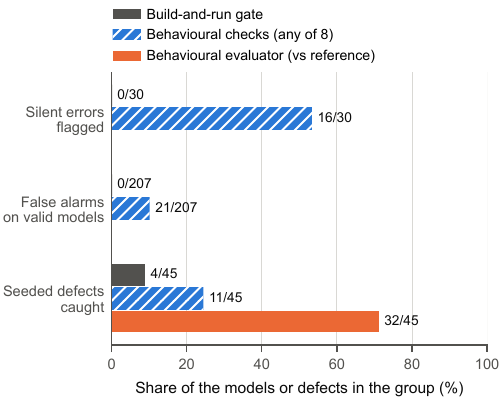}
\caption{Detection by the build-and-run gate, the eight behavioural checks
(a model is flagged if any check fails) and the behavioural evaluator, which
compares the model with its reference: silent errors and valid models of the
main block (243 attempts), and 45 defects seeded into correct models. The
evaluator defines the outcome of the agents' models and is shown for the
seeded defects only.}
\label{fig:detection}
\end{figure}

\paragraph{Detection}
The build-and-run gate blocked none of the 216 runs that had it (all but
C-NoGate), so removing it could not change an outcome. A mutation test explains why: of 45 defects
seeded into correct models it rejected 4, all missing connections that
deadlock a closed line, because wrong routings, distributions, units or
capacities still build and run. We therefore wrote eight behavioural checks
that a gate could run on the model alone --- flow reaches the sink, items are
conserved, Little's law \citep{Little1961} holds, no resource works while the
element it serves is stopped, and four more --- fixed before they were
applied but written knowing the failure modes, so their results are
exploratory. On the 243 models they flagged 16 of the 30 silent errors (53\%
[36, 70]) with false alarms on 21 of the 207 valid models (10\% [7, 15]),
against none for the gate (\cref{fig:detection}). On the 45 seeded defects
they caught 11, all of them also rejected by the gate or by the comparison
with the reference model, which caught 32. The check ``no resource busy while its station is
stopped'' flagged all 15 operator errors of T2.1 without a false alarm; most
false alarms come from Little's law on T3.2, where a large inbound queue gives
an end effect over the short check horizon, and two from T5.2, where few items
reach the sink within it. The sixteenth flag, on a model in the wrong time
unit, is the same end effect of Little's law, not a detection of the unit:
nothing detects H1, H2, H4 or the time units, because these models are internally consistent and answer a different
question, which only a check against the specification can reveal. The
industrial case shows the same kind of error at plant scale
(\cref{sec:case2-runs}), and one in the published expert model itself, found
only because independent models disagreed with it (\cref{sec:case2-val}).

\subsection{RQ3: what the simulator adds over programming from scratch}\label{sec:rq3}

\emph{On six benchmark tasks the same agent writing SimPy was at least as
often valid and far cheaper; what the simulator adds lies in completeness,
realism and inspectability, which the SimPy counterpart of the industrial
case examines at plant scale (\cref{sec:case-simpy}).}

\begin{figure}[htbp]
\centering
\includegraphics[width=\textwidth]{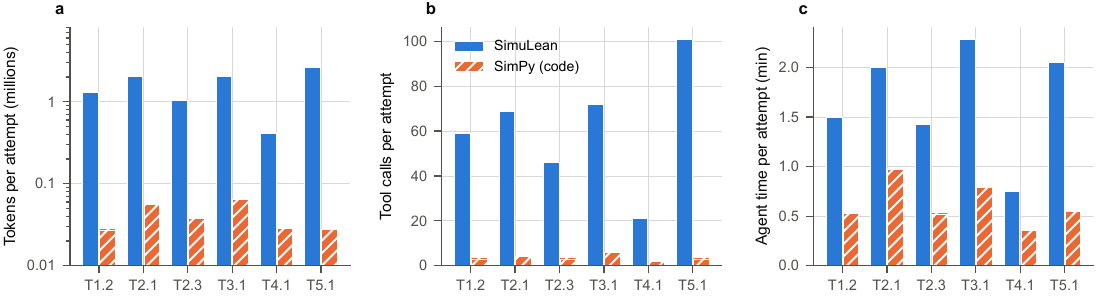}
\caption{The same agent (Claude Opus 5.5) operating \simulean{} (v1, C-Full)
and programming SimPy from scratch on six benchmark tasks, $k=3$: median (a)
tokens per attempt (log scale), (b) tool calls and (c) agent time per
attempt.}
\label{fig:simpy}
\end{figure}

The SimPy arm was valid in 18 of 18 attempts ([0.82, 1.00]) and \simulean{}
v1 in 15 of 18 ([0.61, 0.94]; McNemar $p=0.25$); the three differences are
the T2.1 operator error. The medians were $0.04$ against $1.50\times10^{6}$
tokens, 3 against 62 tool calls and 32 against 109\,s of agent time, and
SimPy needed less on every task (\cref{fig:simpy}). The effort follows the way
of working: the SimPy agent wrote the whole model as one file of about 100
lines, ran it once or twice and reported, whereas the \simulean{} agent
assembled it element by element, each call re-reading a context that holds
the editor's tools and their answers. Three caveats apply: the contract gave
the SimPy agent each KPI's definition, which on T2.1 (``busy working on a
part'') all but rules out counting breaks as busy; the tasks are textbook
sized; and only the SimPy agent had a shell. None of the six tasks needs a 3D
layout, aisle networks with vehicles, calendars, physics, integrated studies
or review by a non-programmer; the industrial case does.

\subsection{RQ4: performance and physical material flows}\label{sec:rq4}

\begin{figure}[htbp]
\centering
\includegraphics[width=\textwidth]{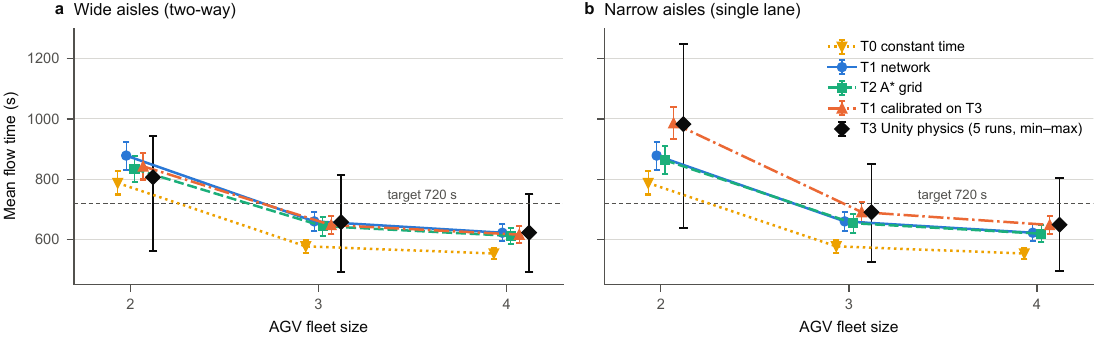}
\caption{Mean flow time of the AGV cell against fleet size for the travel
models, (a) wide two-way aisles and (b) narrow single-lane aisles. Headless
variants: mean of 100 replications with 95\% CI; T3 (Unity navigation and
physics, interactive scene): mean and range of five replications sharing
seeds 1--5 with the headless runs. Dashed: the flow-time target fixed before
the T3 runs.}
\label{fig:physics}
\end{figure}

\emph{The headless engine is fast enough for replicated studies. The
architecture lets physically simulated processes set the durations of the
DES model (demonstrated here on transport), and these reveal effects
that abstract travel models miss, such as congestion in narrow aisles; a
travel model calibrated from them then recovers these effects at headless
cost (\cref{fig:physics}).}

\paragraph{Performance}
On one thread the engine ran the M/M/1 model at 3.0--3.3 million events per
second, tandem lines at 2.4 million (5 stations) to 1.1--1.5 million (50 to
200 stations) and the hold model at 6.9 to 3.3 million; on the same models,
checked to be statistically equivalent, SimPy was 6--14 times and salabim
20--247 times slower per replication. The advantage shrinks on the much
larger model of the industrial case: once the engine is warmed up, a
replication of the agents' models takes 0.1--0.2\,s outside Unity, against
about 0.5\,s for their SimPy models.

\paragraph{Physical material flows}
With wide two-way aisles, physically simulated travel (T3) was within about
2\% of the grid model T2 and of the calibrated network model T1-cal, and every
spatial model chose the same fleet of three AGVs; only constant travel times
(T0) underestimated the flow time, by about 10\% (\cref{fig:physics}a). With
narrow single-lane aisles congestion appeared (\cref{fig:physics}b): at two
AGVs the mean flow time under T3 was 983\,s (five runs, range
638--1\,248\,s) against 878\,s under T1, 863\,s under T2 and 787\,s under T0,
underestimates of 11--20\%, and T1-cal recovered it (986\,s). With the two
targets fixed before the T3 runs (\cref{sec:evaluation}; the flow-time target
is the dashed line) only T0 changed the chosen fleet, through the throughput
target; a flow-time target between the kinematic and the physical value, such
as 900\,s, would change it (an illustration chosen after the results). Physics
is used here as a measuring instrument: a few physical runs (about 190\,s of
computing per simulated hour, in the interactive scene) measure the travel
behaviour, congestion included, and calibrate the travel model, with which the
studies then run headless at full speed (about 1\,ms per simulated hour).

\section{Industrial case: a robotised welding workshop built by one agent}\label{sec:case}

This section shows what one agent does, with no human guidance, with a
request of the size a plant engineer would make (\cref{sec:case-design}). It
reproduces a published industrial study, so that the agent's models can be
checked against an independent model of a real design: the robotised welding
workshop for the nodes of offshore-wind jacket foundations designed with
FlexSim in \citet{LamasRodriguez2020Nodes} and analysed for investment in
\citet{LamasRodriguez2021Investment}. The comparison worked in both
directions: the agent's independent models exposed an error in the
published, expert-built model, and they agree with the corrected model
(\cref{sec:case2-val}).

\subsection{The workshop and the request}\label{sec:case2}
Each jacket needs 3 Y-nodes and 9 X-nodes, each a chord with two stubs. The
workshop cuts and bevels the stubs, fits them on assembly beds with
dimensional controls, makes the root pass, tests it non-destructively (NDT),
welds the node with a robot and tests it again, in an existing building of
five spans served by overhead cranes and two rail carts, on two shifts;
defects send parts back to rework. The agent built scenario~6 of the
published study: 5 assembly workstations, 6 DC2/DC3 workstations, 13
operators, one robot and two rail carts. It received the process and
resources as published and the remaining values (times, probabilities,
capacities, speeds and positions) as data taken from the original model,
which are not reported here. It had to build and verify
the model, run a base case for 28 jackets, an experiment (NDT pairs $\times$
welders) and an optimisation of beds and operators against the takt of one
jacket per week, and report. The specification was written from the
original model and refined, before the runs reported here, wherever a
reading of that model had missed an operating rule; no value was ever
adjusted to match a result. It was used in two forms: the first nine runs
(\cref{sec:case2-runs,sec:case-simpy}) received it with the original model's
crane rule, and the final nine (\cref{sec:case2-val}) with that rule
corrected and two material-handling rules aligned with the running original
model.

\begin{figure}[htbp]
\centering
\includegraphics[width=\textwidth,height=0.62\textheight,keepaspectratio]{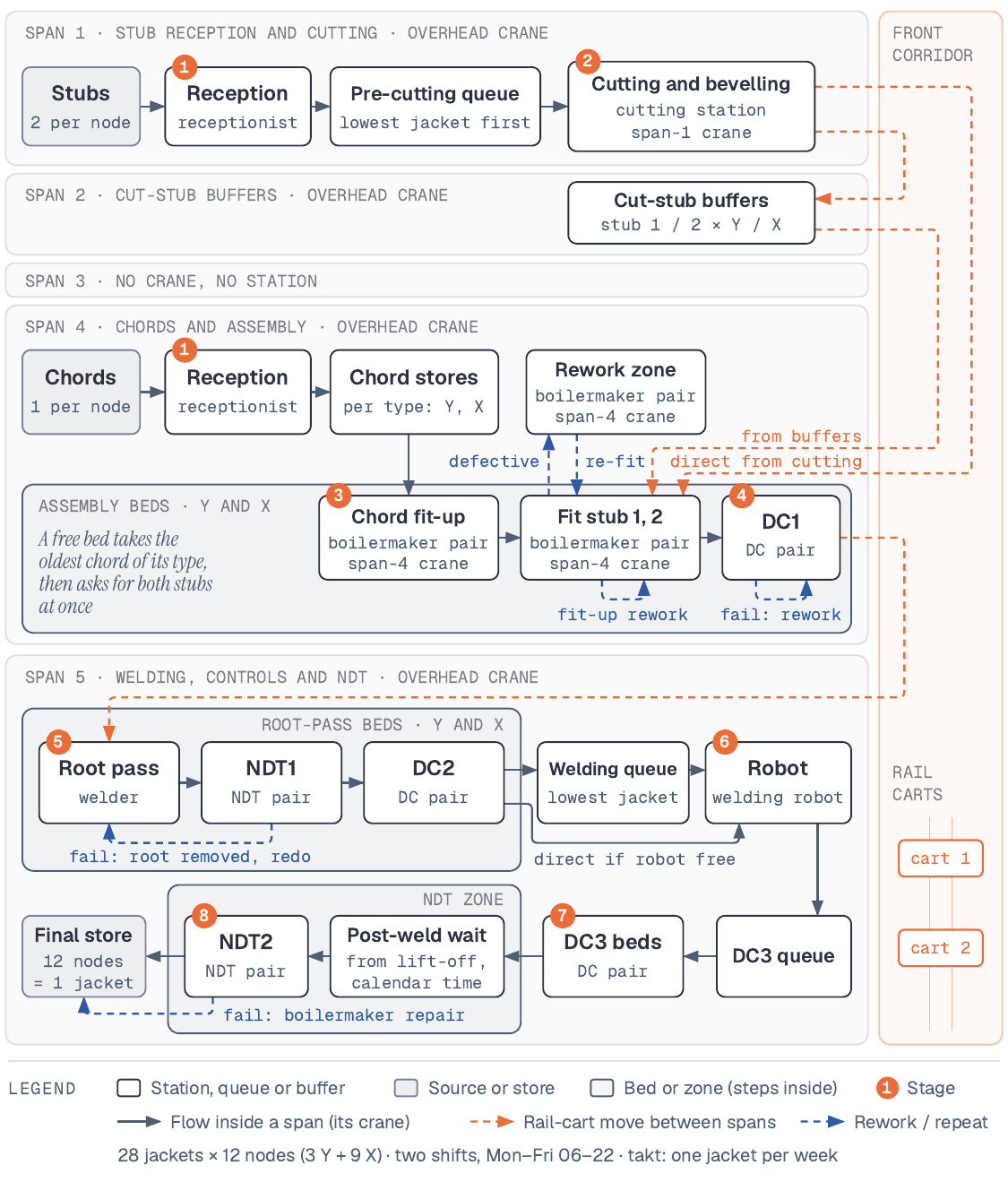}
\caption{Process flow of the robotised welding workshop (scenario~6 of
\citet{LamasRodriguez2020Nodes}), as specified to the agent. Inside a span
its overhead crane moves the parts (span~3 has no crane and no station); two
rail carts in the front corridor carry them between spans. Stubs are cut and
bevelled in span~1 and wait in the buffers of span~2 unless an assembly bed is
waiting for them; chords are stored by node type in span~4. On the assembly
beds the boilermaker pair fits the chord and the two stubs and a
dimensional-control (DC) pair does DC1; in span~5 the welder makes the root
pass, an NDT pair tests it and a DC pair does DC2, then the robot welds the
node, which goes through DC3 and, after a post-weld wait, the final NDT.
Dashed blue: reworks and repeated steps; numbered circles: the stages of the
specification. Times, probabilities, capacities, speeds and positions come
from the original model and are not shown. Diagram drawn with the assistance of Claude (Anthropic).}
\label{fig:case2-flow}
\end{figure}

The request is far larger than a benchmark task (\cref{fig:case2-flow}): a
specification of about 2\,400 words, against about 100 for T1.2; eight
process stages with five rework loops; 13 operators in seven teams, a
receptionist, a cutting station, a welding robot, four overhead cranes that
share their drivers and two rail carts in five spans; and 336 nodes, or
1\,008 parts, per replication. Depending on how each agent structured it, a
\simulean{} model of the corrected specification has 87 to 402 elements, 121
to 512 connections and 21 to 61 reaction rules, and a replication executes 70
to 110 thousand events; a SimPy model has about 700 lines.

\subsection{Six runs on two interface versions}\label{sec:case2-runs}
Three agents worked on v1, the version of the benchmark, and three on v2,
the released version (\cref{sec:harness,sec:case-design}), all on the
first specification. In every run the harness
stopped the first session at its token or tool-call limit and resumed it once
with a fixed message that names the limit and carries no information about
the model; the operator only reopened the agent's scene.
\Cref{tab:case2-runs} gives each run and the means, with the three SimPy runs
of \cref{sec:case-simpy}.

\begin{table}[htbp]
\centering
\caption{The nine runs of the industrial case on the first specification.
Working days per jacket, in real working time: harness re-runs of each
exported model, 50 replications (95\% half-widths 0.007--0.010). Time, calls and cost of the whole run; USD: Claude Code's
API-equivalent estimate. Code: calls of the editor's code-evaluation tool
(\simulean{}) or lines of Python (SimPy).}
\label{tab:case2-runs}
\footnotesize
\setlength{\tabcolsep}{3pt}
\begin{tabularx}{\textwidth}{@{}>{\raggedright\arraybackslash}X c r r r r@{}}
\toprule
Run & Working days & Time (min) & Calls & USD & Code \\
\midrule
v1 rep1    & 4.863 & 97.4  & 244     & 25.1 & 115 \\
v1 rep2    & 4.879 & 130.7 & 2\,756  & 51.7 & 42 \\
v1 rep3    & 4.968 & 82.8  & 1\,281  & 39.0 & 58 \\
v1 mean    & 4.903 & 103.6 & 1\,427  & 38.6 & 72 \\
\midrule
v2 rep1   & 4.915 & 129.4 & 289     & 44.9 & 1 \\
v2 rep2   & 5.125 & 91.9  & 262     & 34.7 & 0 \\
v2 rep3   & 5.065 & 86.3  & 312     & 38.2 & 0 \\
v2 mean   & 5.035 & 102.5 & 288     & 39.3 & 0.3 \\
\midrule
SimPy rep1 & 5.130 & 14.9  & 30      & 2.15 & 973 \\
SimPy rep2 & 5.139 & 14.3  & 40      & 2.46 & 1\,044 \\
SimPy rep3 & 5.111 & 14.1  & 28      & 2.06 & 954 \\
SimPy mean & 5.127 & 14.4  & 33      & 2.22 & 990 \\
\bottomrule
\end{tabularx}
\end{table}

\paragraph{Interface} The v2 agents built, verified, diagnosed, studied,
optimised and presented the workshop without writing code (0, 0 and 1 calls
of the code tool; the one call only checked the open scenes), whereas the v1
agents wrote 42 to 115 blocks of C\# each (148 to 1\,086 lines): loops that create and wire the
stations, their own headless scripts to diagnose stalls, and the optimiser
set up and polled through code. The v2 agents also needed five times fewer
calls, with a much smaller spread (262--312 against 244--2\,756; two v1
agents set fields one call at a time, over a thousand calls each, and hit the
400-call limit in the first session). Effort, however, was the same: building
and verifying the model, up to the start of the experiment, took 53 and
54\,min on average (v1 and v2), with v2's build somewhat cheaper (25 against
30\,USD) and less variable; the whole run took 104 and 103\,min, about 123 and
125\,M cache-read tokens and 39\,USD. The time is set by the tokens the agent
writes and re-reads, not by the number of calls: one batch of 67 edits costs
as much output as the 67 calls it replaces, and almost all tokens are re-reads
of the context.

\paragraph{Model structure} On the first specification the six
\simulean{} models spread by about 5\,\% (4.86--5.12 working days per
jacket); three agents had grouped the beds of a kind into one station with a
count, so that the optimiser could vary it, and their models were the slower
ones. That specification still carried the reference model's routing error
(\cref{sec:case2-val}); on the corrected one all nine agent-built models
agree within 0.8\,\% whatever their structure. The spread was invisible in
the models themselves: every model was self-consistent, its deterministic
workloads matched the original data and its report was plausible --- the
silent errors of \cref{sec:rq2}, at plant scale, which only a comparison with
an independent reference reveals.

\paragraph{Verification} Every agent's own verification found real faults
in its model, two to five per run: a double booking of the rework zone,
cranes held at a full bed, a stub sent to a full buffer, a request left
open. The agents fixed them, the v2 agents with the diagnosis command and
study metrics, the v1 agents with their own scripts.

\begin{figure}[htbp]
\centering
\begin{minipage}[t]{0.49\textwidth}\centering
\includegraphics[width=\linewidth]{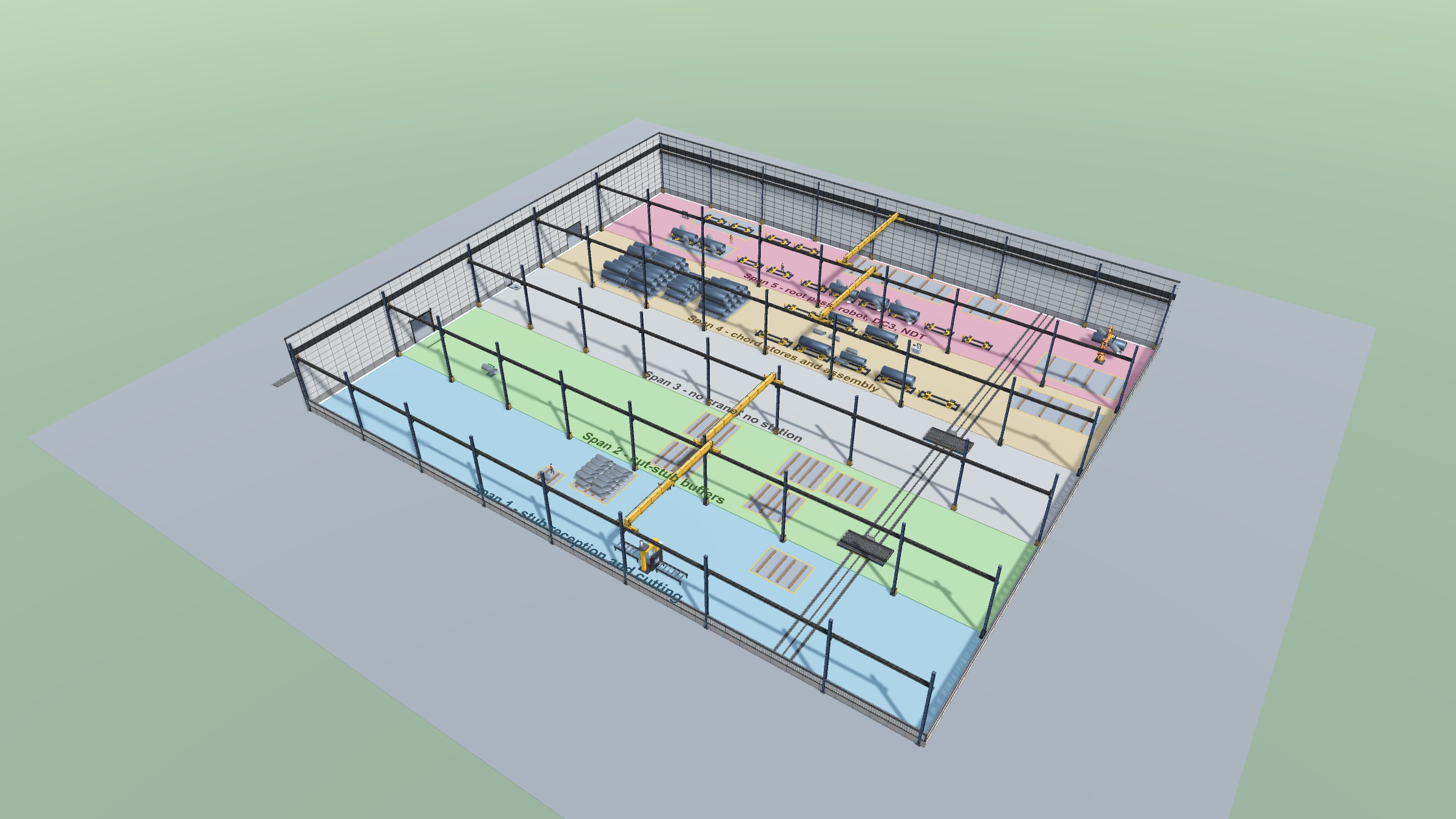}\par\vspace{2pt}
{\footnotesize (a) The five-span workshop}
\end{minipage}\hfill
\begin{minipage}[t]{0.49\textwidth}\centering
\includegraphics[width=\linewidth]{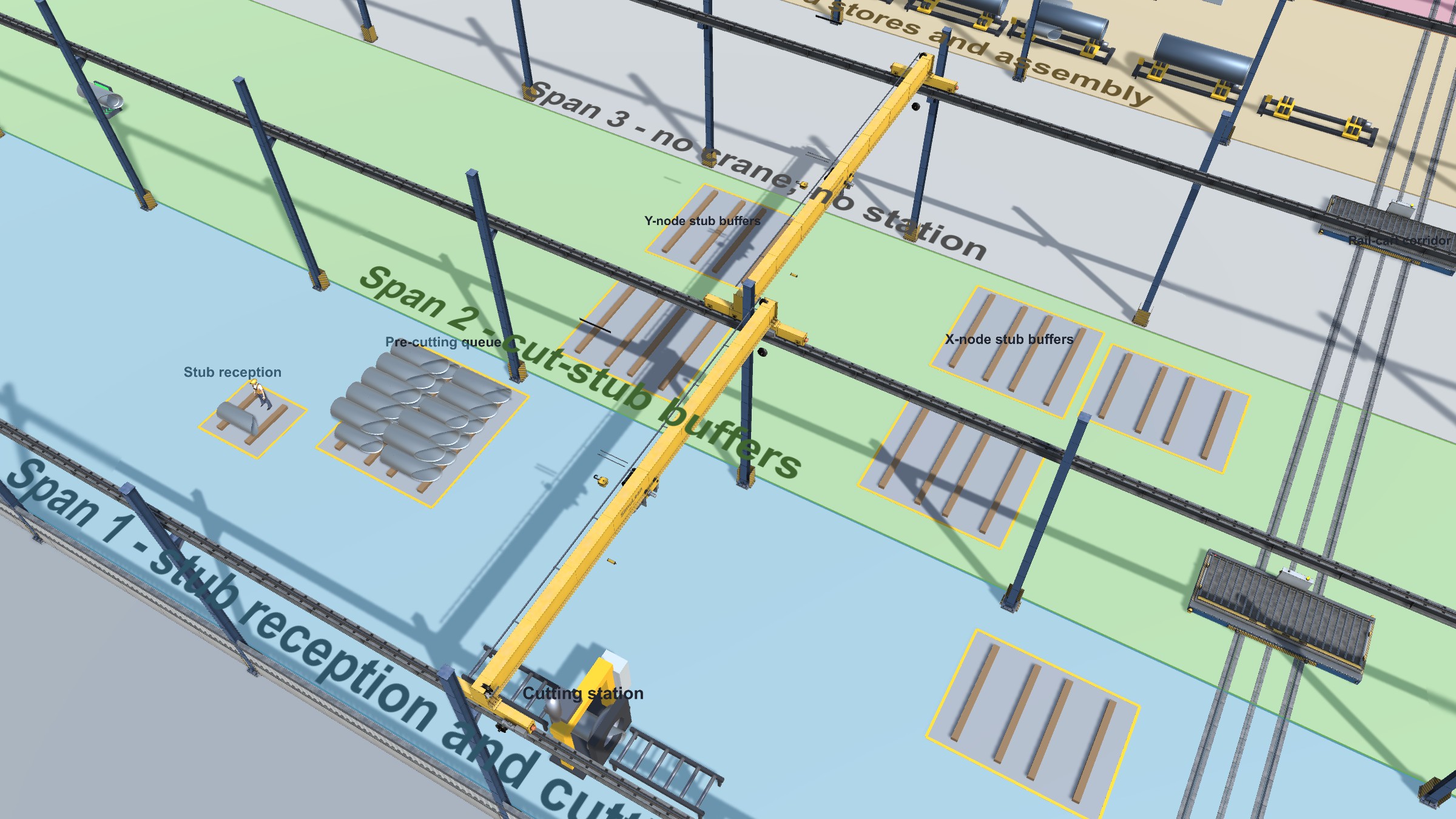}\par\vspace{2pt}
{\footnotesize (b) Stub cutting, buffers and carts}
\end{minipage}\par\vspace{6pt}
\begin{minipage}[t]{0.49\textwidth}\centering
\includegraphics[width=\linewidth]{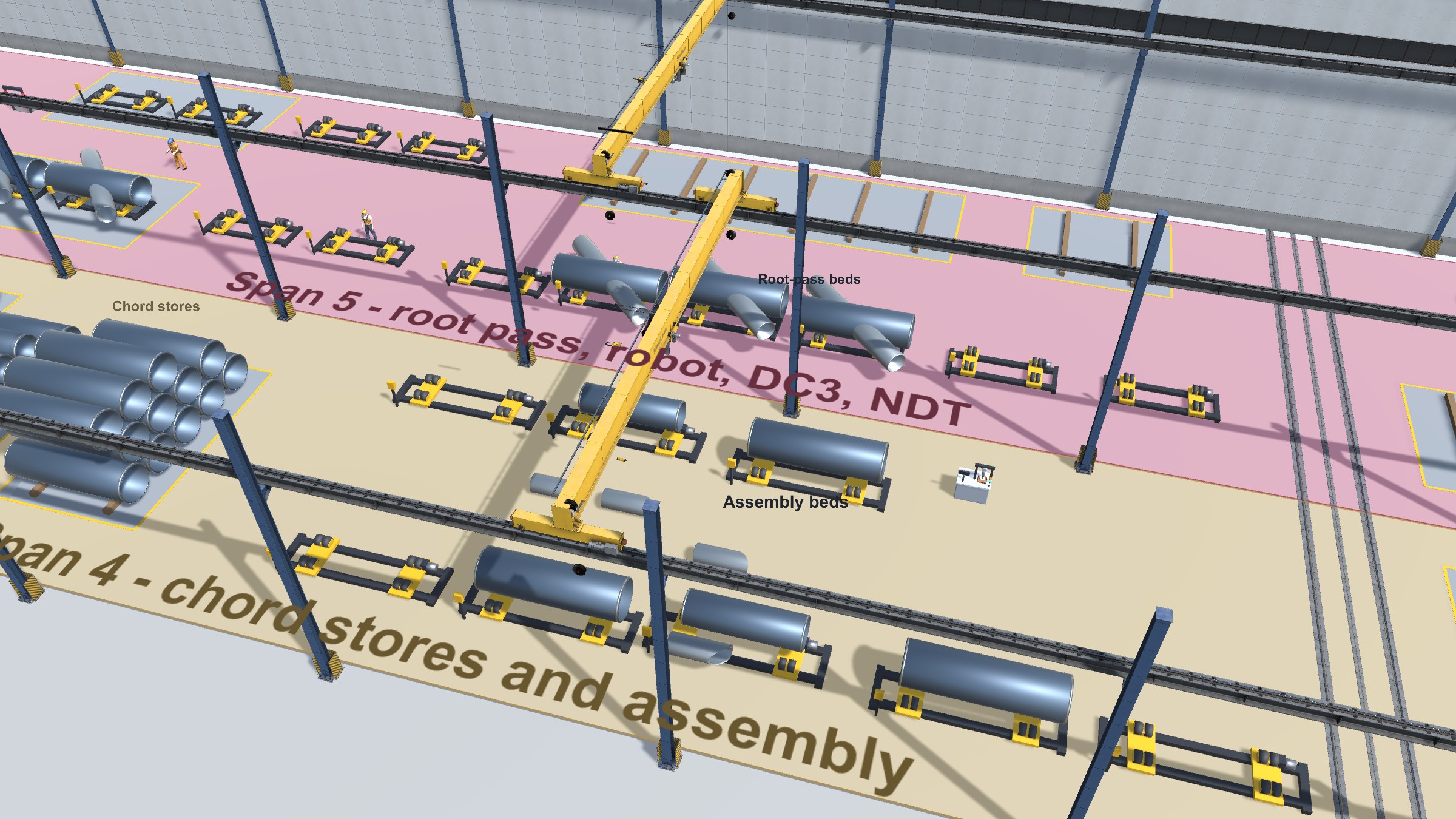}\par\vspace{2pt}
{\footnotesize (c) Assembly and root-pass beds}
\end{minipage}\hfill
\begin{minipage}[t]{0.49\textwidth}\centering
\includegraphics[width=\linewidth]{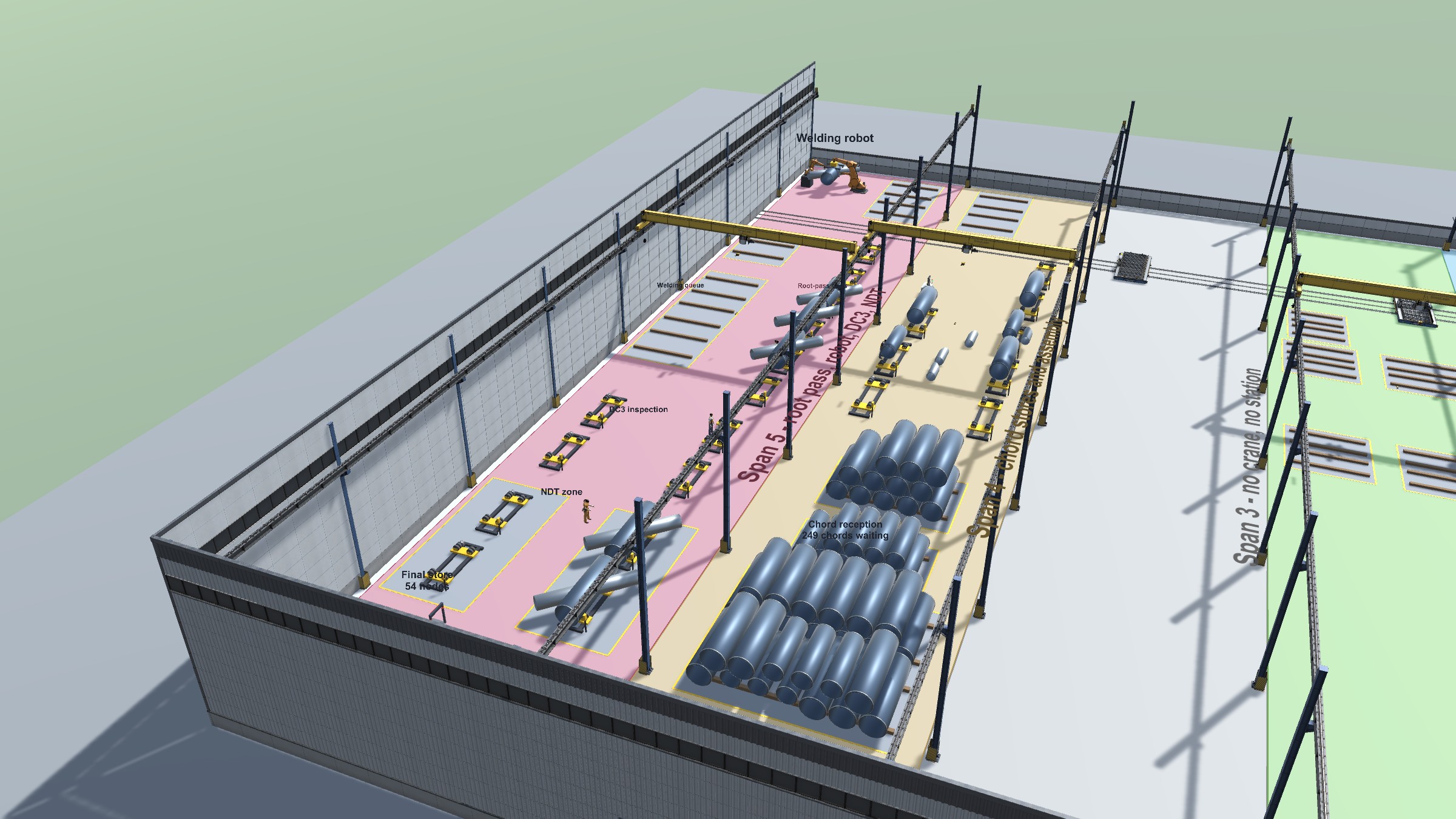}\par\vspace{2pt}
{\footnotesize (d) Robot welding, DC3, NDT and store}
\end{minipage}
\caption{The robotised welding workshop built by the agent of run v1 rep1
(first specification) in \simulean{}. Purely visual work followed the model: a
separate agent session added the building shell, span zones, labels and
cameras, and an interactive agent session with the authors' review (about
half an hour, three rounds) gave the parts and stations generic 3D models
made for this case (tubular chords and stubs, turning-roll positioners, a
pipe-profile cutter, a robot cell) and laid out the waiting parts in piles.
The exported model is unchanged by both (a name-normalised comparison of
the exported configurations shows no change to any element). Views
(b)--(d) were taken while the model was running.}
\label{fig:case2-3d}
\end{figure}

\paragraph{The agents' studies} In run v1 rep1 (\cref{fig:case2-3d}) the base
case gave $4.874 \pm 0.015$ working days per jacket over 20 replications; the
agent identified the bottleneck as the cutting station together with its
span's crane, which the first specification holds during every stub fit --- the rule
later found to be the original model's error (\cref{sec:case2-val}). In the
experiment, with common random numbers, two NDT pairs gave 5.14 working days
per jacket and three 4.88, and a fourth pair or a second welder gained about
0.02 days. The library's genetic algorithm, with five replications per
candidate, found a design with 20 beds and persons instead of 24, at
$4.887 \pm 0.018$ working days over 20 replications.

\subsection{The same workshop programmed in SimPy}\label{sec:case-simpy}
The same agent received the same first specification and limits with the SimPy
contract of \cref{sec:simpy-design}, adapted to the case (the model runs to
the last node and takes the optimisation's counts as parameters; no layout,
library optimiser or 3D). No run hit a limit or needed a resume: each took
about 14\,min and 2\,USD, against about 100\,min and 39\,USD in \simulean{}
(\cref{tab:case2-runs}), and building and verifying took 8\,min against 54.
Each agent wrote about 1\,000 lines of Python, about 700 of them in the
model. The difference lies in the context: a SimPy run re-read 1.4\,M tokens,
a \simulean{} run about 125\,M, since the editor's tools and every answer stay
in the context; and the studies, which run in the editor process
(\cref{sec:arch-overview}), took about 50\,min of each \simulean{} run,
whereas a SimPy replication takes half a second. Outside the editor the
\simulean{} engine runs a replication of the agents' models in 0.1--0.2\,s,
so the command-line runner should cut those studies to a few minutes. The three SimPy models
implement the specification literally and alike --- every bed its own
entity, operators walking back to their waiting point, crane and driver
requested separately --- and their results agree with each other
(5.11--5.14 working days per jacket, against 4.86--5.12 for the six
\simulean{} models). The deterministic workloads match in every model. All
three, however, were slower than the original model (4.84 working days per
jacket) by a similar margin, about 6\,\%, which led to the diagnosis of
\cref{sec:case2-val}; on the corrected specification the SimPy runs gave
4.999 (4.998--5.001) working days per jacket.

For this workshop, code was faster, cheaper and more consistent between
runs. \simulean{} produced a 3D model of the plant that an engineer can see
and question, whereas the SimPy models can only be reviewed by reading their
code, in which the layout exists as coordinates. The command-line runner of
\cref{sec:arch-overview} and a smaller tool context should narrow the cost
gap.

\subsection{The reference model, its correction and the validation}\label{sec:case2-val}
Re-run in FlexSim with 50 replications, the original model reproduces the
published value \citep{LamasRodriguez2020Nodes} and gives
$4.836 \pm 0.008$ working days per jacket (\cref{tab:case2-val}). The three
SimPy models, which follow the specification literally and keep every bed
separate, all came out slower by a similar margin, which a modelling slip of
one agent would not explain. A stage-by-stage diagnosis of the original model
traced the gap to an error in its routing: the stub fit-ups held the crane of
the wrong span. In the corrected model they hold the span-4 (assembly-span)
crane, and the original model gives $4.98 \pm 0.06$ working days per jacket.
Against this corrected reference, the nine agent-built models of the
corrected specification --- three runs each of SimPy and of \simulean{} v1
and v2, each re-run with 50 replications --- are all within $-0.3$ to
$+0.4\,\%$ (\cref{tab:case2-val}), well inside the reference's own
uncertainty. The deterministic
workloads --- the busy hours of the welder, the robot, the cutting station and
the NDT pairs --- matched the original data in every model. With the
correction, scenario~6 sits at the takt limit of one jacket per week, at about
5.0 working days per jacket (4.96--5.00 across the models), met in about half
of the replications. On the corrected specification a run took 12\,min, 19 calls and
1.9\,USD on average in SimPy (building and verifying 8\,min, 1.6\,USD),
134\,min, 185 calls and 20\,USD in \simulean{} v1 (39\,min, 16\,USD) and
112\,min, 339 calls and 37\,USD in v2 (52\,min, 26\,USD); every \simulean{}
run hit the token limit once and was resumed, no SimPy run hit a limit. The v1
agents built this time with fewer, larger C\# blocks that call the commands in
loops --- 60 to 69 blocks and 1\,038 to 1\,168 lines per run, as much code as
the SimPy agents wrote --- so v1 was cheaper than v2 in this round, whereas the
v2 agents again wrote no code (161 to 299 domain commands per run). Both
interface versions give valid models, and code written from scratch is
10--20 times cheaper.

\begin{table}[htbp]
\centering
\caption{The reference model, its correction and the agent-built models of
the corrected specification, in working days per jacket (real working time).
Mean $\pm$ 95\% half-width over 50 replications, or mean (range) over three
runs, each re-run with 50 replications. FlexSim reports the original study's
KPI; its results are converted to working days with the model's working
calendar (as run, the original model reproduces the published value). In
every model the busy hours of the welder, the robot, the cutting station and
the NDT pairs equal the workloads implied by the original data.}
\label{tab:case2-val}
\footnotesize
\begin{tabularx}{\textwidth}{@{}>{\raggedright\arraybackslash}X >{\centering\arraybackslash}p{3.2cm} >{\centering\arraybackslash}p{3.4cm}@{}}
\toprule
Model & Working days per jacket & Difference from corrected \\
\midrule
Original as run, with the routing error (FlexSim) & $4.836 \pm 0.008$ & $-2.9\%$ \\
Original corrected (FlexSim) & $4.98 \pm 0.06$ & reference \\
Agent-built, SimPy, 3 runs & 4.999 (4.998--5.001) & $+0.4\%$ \\
Agent-built, \simulean{} v1, 3 runs & 4.986 (4.964--4.998) & $+0.1\%$ ($-0.3$ to $+0.4\%$) \\
Agent-built, \simulean{} v2, 3 runs & 4.991 (4.983--4.999) & $+0.2\%$ ($+0.1$ to $+0.4\%$) \\
\bottomrule
\end{tabularx}
\end{table}

The usual direction of verification is thus reversed: an independent model,
built by an agent from the text, served as a behavioural oracle for the
expert's model. The first author, who prepared the specification, was among
the authors of the original model; an error of this kind, found and fixed
when a model is rebuilt independently, is normal modelling practice. It also
shows that agreement with a published value is not by itself evidence of
validity: before the correction, the \simulean{} models closest to the
original model were ones whose agents had declared approximations of some
rules.

\section{Discussion}\label{sec:discussion}

\subsection{Implications}\label{sec:implications}
\paragraph{One agent builds complete models that people can inspect} A
valid benchmark model took about two minutes of agent time and about one
dollar of estimated API cost, and the industrial workshop --- five spans,
overhead cranes and rail carts, two shifts, rework loops and 13 operators ---
was built and verified in under an hour on average and fully studied in 83
to 185\,min per run. A panel of simulation researchers recently expected
generative AI to generate valid models autonomously only within the next two
decades \citep{Ehm2025LookFuture}; for the model-building part of a study,
and for systems of this size, our results suggest that this capability is
already within reach. The result is not a script
but a 3D scene with the plant's layout and the request's names, which an
engineer can watch, edit and hand back to the agent. For small and
medium-sized manufacturers, which lack simulation specialists more than
computing power, that may matter more than the cost of the agent's tokens.

\paragraph{A verified platform or unverified code} For a frontier agent,
code is the cheaper medium: about one thirtieth of the tokens on six benchmark
tasks, at equal or higher validity, so the capability is not tied to one
tool. What an agent writes from scratch, however, contains only what it
thought of, and its semantics are tested by nobody but the agent. A simulator
supplies such features as configuration whose semantics are tested once,
against reference models and analytical results, for every model built with
it. The industrial case confirms the trade-off at plant scale
(\cref{sec:case-simpy}): in SimPy the agent was faster (11--15 against
83--185\,min), cheaper (about 2 against 20--50\,USD) and more consistent between runs,
and wrote about 1\,000 lines with no layout; in \simulean{} it produced a 3D
model that an engineer can see and question. Efficiency favours code;
review and communication favour the simulator. In a design project the model
is what the engineer checks and what the client looks at before committing to
the investment, and against the cost of a modelling project the 20--50\,USD
of an agent-built 3D model is small; the command-line runner and a smaller
tool context should narrow the gap further. Which medium suits a project
therefore depends on whether it needs a quick numerical answer or a model
that people other than its author must understand and trust.

\paragraph{Physics to measure, DES to study} The engine lets a physically
simulated process end a DES duration: physics reveals effects that abstract
models miss, such as congestion in narrow aisles, and calibrates the fast
headless model on which the studies run. Shown here on transport, the route
``calibrate with physics, study headless'' extends to handling and placing
material and beyond manufacturing, for example to fleets of unmanned surface
vessels whose travel times depend on vessel dynamics.

\paragraph{What the interface changes: traceability, not validity} None
of the interface interventions changed validity: v0 and v1, the ablations and
the two versions in the industrial case gave the same validity. We read this as a ceiling rather than as evidence that the design
does not matter: \simulean{} had been developed with AI coding agents from
the start, so v0 was already usable by them, and the silent errors that
remained were errors in reading the specification, not in operating the tool.
The one operating problem, agents polling a long study until the token cap
(four runs at T4.2), came from a command v1 lacked and the released version
adds: a wait on a running study. What
the interface did change is how the model is built. Where the commands cover
the task, the agents used nothing else --- 105 of the 108 C-Full attempts on v1
and every v2 run of the industrial case wrote no model code --- and the model comes
with a record of every step in the modelling vocabulary; where they did not,
as for bulk edits, diagnosis and the optimiser on v1, the agents fell back on
as much code as a SimPy model (about 1\,100 lines per run in the case). For
verification this is the difference that matters: a model assembled from
verified elements through recorded commands can be reviewed as a
configuration and a 3D scene, whereas a model written as code is new software
to be reviewed line by line every time. Reviewing a configuration is not
free --- the case models have 87 to 402 elements and 21 to 61 reaction rules,
and agents structured the same workshop differently --- but what has to be
checked is the structure and the parameters, not the semantics. Commands do
not reduce effort, which is set by the tokens an agent writes and re-reads
(about 99\% of them re-reads of its context); that needs smaller contexts and
answers.

\paragraph{Verification has to be behavioural} Every failure we observed was
silent, so trust in an agent-built model cannot rest on the fact that it
builds and runs: the build-and-run gate blocked no agent model and rejected 4
of 45 seeded defects. Checks on the behaviour of the model flagged 16 of 30
silent errors, an upper bound since they were written knowing the failures.
At plant scale the same holds: all workshop models were self-consistent,
with exact workloads and plausible reports, yet on the first specification
their results spread by about 5\,\% for reasons that nothing in the models
showed. The oracle also works in reverse: three literal SimPy models that
agreed with each other but not with the published model led to the discovery
of an error in that expert-built model, and against the corrected model the agent-built
models agree within 1\%. An independent model --- a published
one, one in another tool or one by another agent --- is therefore the oracle
that matters, for the agent's model and the expert's alike, and at the cost
of an agent-built model, building a model twice is an affordable practice for
decisions that matter. A simulator built for
agents should offer behavioural checks, reproducible runs and a record of the
agent's assumptions as much as it offers commands.

\paragraph{Cost and time against a modeller} The industrial workshop --- a
complete, verified 3D model with an experiment and an optimisation --- took
one and a half to three hours of agent time and 20--50\,USD per run in
\simulean{}.
From their experience, the authors estimate that a simulation engineer would
need about two weeks of dedicated work, including the data analysis, to build
and debug a model like this, without its 3D presentation, at a much higher
salary cost; this is an estimate, not a measurement. The human role shifts to specifying, checking
and interpreting.

\subsection{Where agents fail}\label{sec:disc-failures}
Vague or contradictory specifications produce silent errors that no check on
the model alone can find: in H4 every attempt doubled an already doubled
demand, in H1 agents added a rule, and on T4.3 two attempts read the task's
minutes as hours. The second model failed more often on the harder tasks, but
in the same ways. The agents asked no questions; they
assumed and reported their assumptions. Asking the agent to list a request's
ambiguities before building is a cheap remedy we did not test. Where a
concept is missing, agents improvise: the dominant benchmark error, an
operator kept busy during a break, arises because the break is attached to
the station, and calendars for resource teams would remove the trap. At plant
scale agents also simplify: to let the optimiser vary bed counts, three of
six grouped the beds of a kind into one station. On the first specification
these models were about 3\,\% slower, a difference that vanished once the
reference was corrected but that nothing in the models revealed. A model's
structure should therefore be checked against the specification, not only
its parameters and behaviour.

\subsection{Limitations}\label{sec:limitations}
\begin{itemize}[leftmargin=*,itemsep=2pt]
\item \emph{Author-built tasks.} The authors wrote the tasks, reference
models and criteria, fixed before the runs; the SimPy arm, the behavioural
checks and the case came after the first results. The case's reference model
was published before this work, but the first author was among its authors
and prepared the specification; it was corrected during this work, and its
final specification was run three times per tool.
\item \emph{One industrial case.} The plant-scale evidence rests on one
workshop and one complete specification; other plants, and requests less
complete than a published study, may give other results.
\item \emph{One provider, $k=3$.} Two models of one provider: only large
differences between versions or ablations are detectable. The simulator was
developed with the help of AI coding agents of the same provider, which may
have adapted its interface to them; other agents may do worse.
\item \emph{No simulator built for people as a baseline.} We compared
\simulean{} with code written from scratch, not with a simulator that was not
designed for agents (e.g.\ a commercial package driven through a scripting or
MCP interface); the value of the design itself is therefore argued from
traceability and from the ceiling of the interface interventions, not
measured against such a baseline.
\item \emph{Physics is a model too.} T3 is Unity's navigation and physics,
not measured vehicle behaviour, and the narrow-layout T3 runs are few.
\item \emph{Studies in the editor.} The case's studies ran inside the editor,
about 10--14 times slower than outside it, which inflates the time of the
\simulean{} runs.
\item \emph{The SimPy comparison.} Its contract defined every KPI and only
the SimPy agent had a shell.
\item \emph{Commercial tools not measured.} \Cref{tab:capabilities} draws on
published descriptions.
\item \emph{Verification, not validation.} Validity is equivalence with a
reference or published model; validation against the real system remains the
modeller's task \citep{Law2024}.
\end{itemize}

\section{Conclusions and future work}\label{sec:conclusions}

We presented \simulean{}, a free and open-source 3D discrete-event simulator
designed to be programmed by AI agents, and evaluated it with a benchmark
whose validity criteria were fixed before the runs, a comparison with code
written from scratch, an industrial case checked against a published model
that this work corrected, and performance and physics experiments.

\textbf{RQ1.} One frontier agent working alone built valid models in 93 of
108 attempts (86\%; 77 of 84 on the main tasks), with a median of 2.0\,min and
an estimated 1.14\,USD per valid model; a second model was valid in 43 of 54
attempts over the same tasks (Opus 46 of 54, no significant difference). At plant scale, one agent rebuilt a published robotised
welding workshop, verified it and ran its experiment and optimisation in one
and a half to three hours and for 20--50\,USD per run; the comparison exposed an
error in the published model, and against the corrected model the
agent-built models agree within 1\%.
\textbf{RQ2.} Every remaining failure was a silent error: the build-and-run
gate blocked none of 216 gated runs, behavioural checks flagged 16 of 30 with
21 false alarms in 207 valid models, and errors in reading the specification
escaped every check on the model. In the industrial case, the agents'
models disagreed by about 5\,\% on the first specification for reasons
invisible in the models, and independent agent-built models revealed an
error in the published expert model itself.
\textbf{RQ3.} A trade-off: the agent writing SimPy was at least as valid (18
of 18 against 15 of 18) and used about one thirtieth of the tokens, while
\simulean{} adds what the code left out or simplified --- aisle networks,
cranes, calendars, failures and teams as configuration, integrated studies,
physics calibrated for headless runs, and a model a non-programmer can see,
review and edit. The industrial case confirmed it: in SimPy the agent was
faster (about 13 against 100--135\,min on average), cheaper (about 2 against
20--40\,USD) and more consistent; in \simulean{} it produced a 3D model an
engineer can question. What the commands add is traceability rather than
validity: where they cover the task the agents wrote no code, and the model
is a configuration of verified elements with a record of how it was built;
where they did not, the agents wrote as much code as in SimPy.
\textbf{RQ4.} The engine runs one to three million events per second outside
Unity, 6--14 times faster than SimPy on the benchmark's models and about three
times faster on the larger model of the industrial case; physically simulated transport reveals
congestion that abstract travel models miss, and a travel model calibrated
from it recovers the results at headless cost.

\paragraph{What is provided} \simulean{} is released under the GPL-3.0 as a
free, open and capable 3D DES tool, designed for agents and equally usable by
people. It is meant for teaching as well as for industrial projects. We hope that lecturers will find in it a free
alternative to commercial packages for their courses, one in which students
can build, watch and question a model with or without an agent. We will keep
developing it on such cases and welcome contributions from the community;
under its licence, anyone may also adapt a version of their own.

\paragraph{Future work} Development continues on new features and on real
industrial cases; on libraries with presets for specific industries; on the
command-line runner, so that studies and the optimiser run outside the
editor; and on a smaller tool context, to make the simulator more efficient
for agents. We will also keep benchmarking it, with other agents and stronger
behavioural checks, and later with open-weight models running locally, which
keep plant data on premises but become practical only once the context
shrinks: about 99\% of the tokens of an agent-built model are re-reads of its
context (about 125\,M per run in the industrial case).

\section*{CRediT authorship contribution statement}
\textbf{Javier Pernas-\'Alvarez} and \textbf{Diego Crespo-Pereira}
contributed equally to all aspects of this work: Conceptualization,
Methodology, Software, Validation, Formal analysis, Investigation, Data
curation, Resources, Visualization, Writing -- original draft, Writing -- review \&
editing, Supervision and Project administration.

\section*{Declaration of competing interest}
The authors declare that they have no known competing financial interests or
personal relationships that could have appeared to influence the work reported
in this paper.

\section*{Funding}
This research did not receive any specific grant from funding agencies in the
public, commercial, or not-for-profit sectors.

\section*{Data and code availability}
\simulean{} is released under the GNU GPL-3.0 at
\url{https://github.com/simulean/SimuLean}; version 1.0.0, the one described here, is archived on Zenodo \citep{SimuLeanZenodo}. The benchmark tasks, reference models, agent transcripts and
analysis scripts are available from the authors on request.

\section*{Declaration of generative AI and AI-assisted technologies in the manuscript preparation process}
During the preparation of this work the authors used Claude (Anthropic),
through the Claude Code tool, to help draft and edit the text, search and
summarise the literature, write the analysis scripts, produce the data
figures from the experimental data and draw the diagrams (Figs.~\ref{fig:designed-for-agents},~\ref{fig:positioning},~\ref{fig:architecture},~\ref{fig:build-path},~\ref{fig:design} and~\ref{fig:case2-flow}). After using this tool, the authors reviewed and edited the content as
needed, checked every reported number against the underlying data, and take
full responsibility for the content of the published article.

\paragraph{AI assistance in the development of the software} Parts of
\simulean{} --- source code, tests and documentation --- were written with
the assistance of AI coding agents (Claude Code with Claude models,
Anthropic) working under the authors' direction. Every change passed the
simulator's automated test suite and reference models
(\cref{sec:simulean}), and the authors reviewed and are responsible for the
released code.

\paragraph{AI agents as the object of the study} The AI agents that build the
simulation models in the experiments (Claude Opus 5.5 and Claude Sonnet 5.5,
through Claude Code) are the subject of the research, not a writing aid; their
use is described in \cref{sec:evaluation}.

\bibliographystyle{elsarticle-num-names}
\bibliography{references}

\end{document}